\documentclass[10pt,twocolumn,letterpaper]{article}

\usepackage{cvpr}              % To produce the CAMERA-READY version
\usepackage{xspace}
\usepackage{amsmath}
\usepackage{algorithm}
\usepackage{algorithmic}
\usepackage{multirow}
\usepackage{makecell}
\usepackage{pifont}
\usepackage{color}
\usepackage{colortbl}

\def\onedot{.\xspace}
\def\eg{\emph{e.g}\onedot} 

\def\ie{\emph{i.e}\onedot}

\newcommand{\cmark}{\ding{52}\xspace}%
\newcommand{\xmark}{\ding{56}\xspace}%

\definecolor{blue_tab}{RGB}{227, 240, 251}
\definecolor{cvprblue}{rgb}{0.21,0.49,0.74}
\usepackage[pagebackref,breaklinks,colorlinks,citecolor=cvprblue]{hyperref}

\def\paperID{*****} % *** Enter the Paper ID here
\def\confName{CVPR}
\def\confYear{2026}

\newcommand{\gbb}[1]{\textcolor{red}{#1}}  % mark verion Bin-Bin Gao
\renewcommand{\gbb}[1]{#1} % clean version Bin-Bin Gao
\def\method{Real-IAD Variety}

\title{\method: \\ Pushing Industrial Anomaly Detection Dataset to a Modern Era}

\author {Wenbing Zhu\textsuperscript{a,d}\thanks{Equal contribution}, 
\quad Chengjie Wang\textsuperscript{b,c}$^{*}$,
\quad Bin-Bin Gao\textsuperscript{b}$^{*}$, \\
\quad Jiangning Zhang\textsuperscript{b}, 
\quad Guannan Jiang\textsuperscript{e}, 
\quad Jie Hu\textsuperscript{f}, 
\quad Zhenye Gan\textsuperscript{b}, 
\quad Lidong Wang\textsuperscript{a}, 
\quad Ziqing Zhou\textsuperscript{a}, \\
\quad Jianghui Zhang\textsuperscript{g},
\quad Linjie Cheng\textsuperscript{a}, 
\quad Yurui Pan\textsuperscript{a}, 
\quad Bo Peng\textsuperscript{g}, 
\quad Mingmin Chi\textsuperscript{a}\thanks{Corresponding authors.}, 
\quad Lizhuang Ma\textsuperscript{c} \\
\normalsize 
\textsuperscript{a}{Fudan University} \quad 
\textsuperscript{b}{Youtu Lab, Tencent} \quad  
\textsuperscript{c}{Shanghai Jiao Tong University} \quad 
\textsuperscript{d}{Rongcheer Co., Ltd} \quad \\ 
\normalsize 
\textsuperscript{e}{City University of Hong Kong}
\textsuperscript{f}{National University of Singapore}
\textsuperscript{g}{Shanghai Ocean University} \\
{\tt\small Louis.zhu@rongcheer.com, jasoncjwang@tencent.com, csgaobb@gmail.com}\\
{\tt\small \{vtzhang,wingzygan\}@tencent.com, gujiang@um.cityu.edu.hk, hujie.cpp@gmail.com} \\
{\tt\small m240751960@st.shou.edu.cn, \{ldwang23, zqzhou23, ljcheng24, yrpan24\}@m.fudan.edu.cn} \\
{\tt\small bpeng@shou.edu.cn, mmchi@fudan.edu.cn, ma-lz@cs.sjtu.edu.cn} \\
{\tt\small Website: \url{https://realiad4ad.github.io/Real-IAD-Variety}}
}

\begin{document}
\maketitle

\begin{abstract}

\gbb{Industrial Anomaly Detection (IAD) is a cornerstone for ensuring operational safety, maintaining product quality, and optimizing manufacturing efficiency. However, the advancement of IAD algorithms is severely hindered by the limitations of existing public benchmarks. Current datasets often suffer from restricted category diversity and insufficient scale, leading to performance saturation and poor model transferability in complex, real-world scenarios. To bridge this gap, we introduce Real-IAD Variety, the largest and most diverse IAD benchmark. It comprises 198,950 high-resolution images across 160 distinct object categories. The dataset ensures unprecedented diversity by covering 28 industries, 24 material types, 22 color variations, and 27 defect types. Our extensive experimental analysis highlights the substantial challenges posed by this benchmark: state-of-the-art multi-class unsupervised anomaly detection methods suffer significant performance degradation (ranging from 10\% to 20\%) when scaled from 30 to 160 categories. Conversely, we demonstrate that zero-shot and few-shot IAD models exhibit remarkable robustness to category scale-up, maintaining consistent performance and significantly enhancing generalization across diverse industrial contexts. This unprecedented scale positions Real-IAD Variety as an essential resource for training and evaluating next-generation foundation IAD models.} 
\end{abstract}
\section{Introduction} \label{sec:intro}

The field of anomaly detection has undergone a paradigm shift towards fully unsupervised~\cite{softpatch,softpatch+,m3dm}, multi-class unsupervised (unified)~\cite{uniad,gao2024onenip,guo2025dinomaly}, and zero-/few-shot~\cite{iclr2024anomalyclip,gao2025adaptclip,metauas} learning frameworks, \gbb{ushering in a transformative era for model architecture.}
%marking a transformative era in model design.
Within the specialized domain of Industrial Anomaly Detection (IAD), substantial progress has been achieved in addressing the inherent complexity and variability of industrial processes.

\begin{figure*}[ht]
\centering
\includegraphics[width=1.0\linewidth]{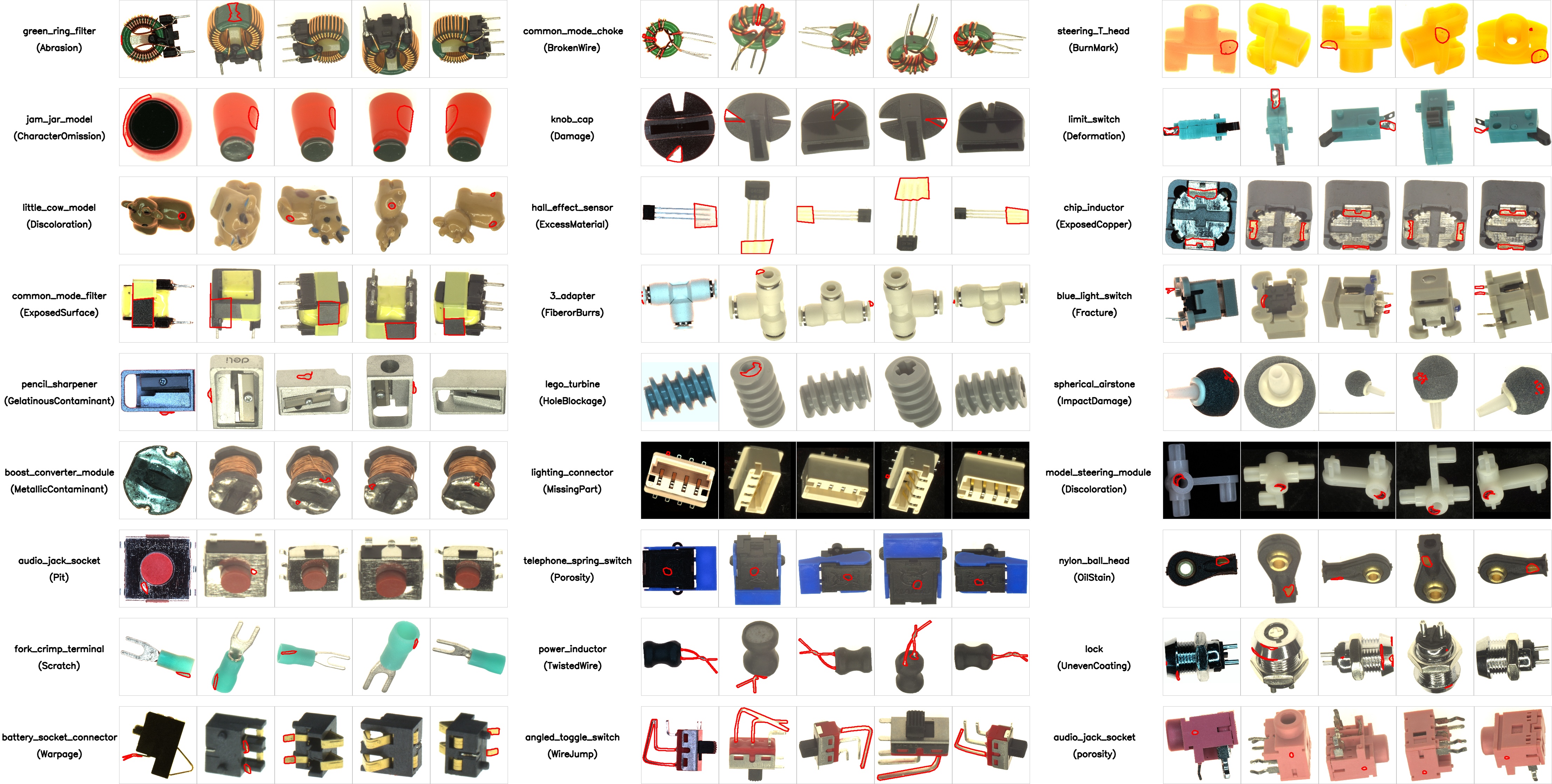}
\caption{
\gbb{
Visual illustration of diverse anomaly patterns in the Real-IAD Variety dataset. We present representative samples across 27 defect types, where anomaly regions are highlighted with red boundaries. Each sample is captured from five distinct viewpoints, demonstrating that certain defects are only detectable from specific angles while remaining invisible in others. Text annotations follow the format of ``object (defect type)", covering both structural damages (\eg, scratch) and logical inconsistencies (\eg, missing part).
}
}
\label{fig:typical_samples}
\end{figure*}

\begin{figure}[t]
\centering
\includegraphics[width=0.9\linewidth]{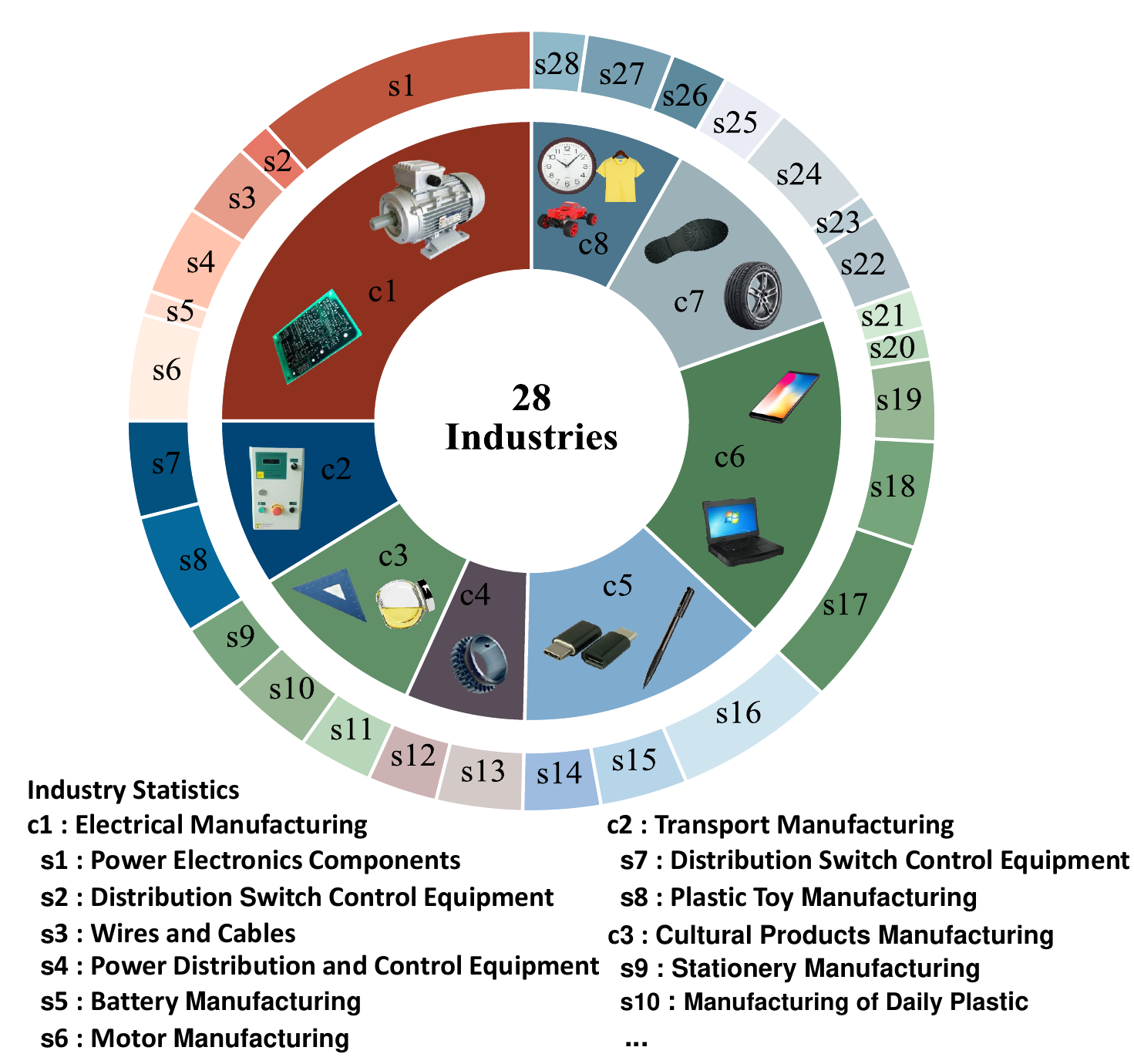}
\caption{
\textbf{Industry distribution of the proposed Real-IAD Variety dataset.}
The dataset encompasses 8 major industrial groups (denoted as \textit{c}): electrical, transport, cultural products, metal, general, electronics, rubber plastic, and other manufacturing sectors.
These major categories are further subdivided into 28 industrial subcategories (denoted as \textit{s}), and
complete details are provided in Appendix.
}
\label{fig:teaser}
\vspace{-1em}
\end{figure}
Recent IAD research has predominantly concentrated on developing unified models capable of detecting anomalies across multiple industrial objects using a single model~\cite{uniad,cfa,simplenet,mambaad,gao2024onenip}. Concurrently, zero-shot and few-shot anomaly detection approaches~\cite{anomalygpt,iclr2024anomalyclip,gao2025adaptclip,metauas} have emerged to identify anomalies in previously unseen domains without requiring \gbb{task-specific fine-tuning}. These advances are fundamentally reshaping IAD enabling more robust and adaptable frameworks. \gbb{However, the progress is increasingly bottlenecked by the limitations of existing benchmarks.}

Despite the contributions of \gbb{established} benchmarks such as MVTec~\cite{mvtec}, VisA~\cite{visa}, PAD~\cite{pad}, and Real-IAD~\cite{realiad}, these benchmarks exhibit critical limitations in category diversity and scenario coverage.
\gbb{For instance}, Real-IAD~\cite{realiad}, offers the largest category count, \gbb{is limited to 30 categories}, while the widely adopted MVTec and VisA datasets comprise only 15 and 12 categories, respectively. 
Therefore, it constrains the comprehensive evaluation of contemporary unified and zero-/few-shot IAD models.
Furthermore, while synthetic data generation via pre-trained diffusion models has been explored~\cite{anogen,anomalyany,seas}, these approaches still face challenges in generating authentic and diverse anomalous patterns.

\gbb{To bridge this gap and propel IAD research} toward unified and zero-/few-shot paradigms, \gbb{there is an urgent requirement for benchmarks} that encompass substantially broader industrial categories and defect typologies.
In this paper, we introduce Real-IAD Variety, a large-scale benchmark that represents a significant advancement in the field.
As illustrated in Figs.~\ref{fig:typical_samples}, ~\ref{fig:teaser} and~\ref{fig:statistic}, this dataset substantially expands the scope of IAD data collection, featuring 160 categories spanning 28 industries, 24 material types, 22 color variations and {\gbb{27 defect types}, \gbb{culminating in} a total of \gbb{198,950} images with high-fidelity pixel-level annotations.
The category count of Real-IAD Variety is approximately 5.3 times that of \gbb{its predecessor}, Real-IAD~\cite{realiad}, making it the first truly multi-industry benchmark that overcomes the constraints of prior datasets for unified and zero-/few-shot evaluation.

\gbb{Beyond scale}, Real-IAD Variety offers unprecedented diversity in materials (\eg, composites, metals, ceramics) and industries (\eg, electrical manufacturing, transport, electronics), addressing the practical needs of real-world deployment. We leverage this benchmark to establish rigorous evaluations across three critical settings: multi-class unsupervised anomaly detection (MUAD), multi-view anomaly detection, and zero-/few-shot anomaly detection. 

\gbb{Our extensive experiments yield two pivotal insights, \textbf{scalability bottlenecks} and \textbf{paradigm robustness}. We observe that an increase in category diversity leads to a significant performance degradation (10\%-20\%) in state-of-the-art MUAD methods, revealing their limited scalability when tasked with handling multiple category distributions. In contrast, zero-/few-shot approaches, particularly those leveraging large-scale vision-language models, demonstrate remarkable robustness. Their performance remains consistent regardless of category count, eventually surpassing MUAD methods once a certain complexity threshold is reached.}

In summary, our contributions are threefold:
\begin{itemize}
\item We present Real-IAD Variety, a large-scale IAD dataset featuring unprecedented category diversity, extensive industry coverage, and high-quality pixel-level annotations. It expands the category count to 160 across 28 industries, 24 material types, 22 color variations and \gbb{27 defect types} with \gbb{198,950} images, representing a substantial advancement over existing IAD benchmarks.

\item We establish comprehensive benchmarks on Real-IAD Variety across three critical IAD settings: multi-class unsupervised anomaly detection, multi-view anomaly detection, and zero-/few-shot anomaly detection, facilitating systematic evaluation of current methodologies.

\item We reveal that increasing category count significantly impairs the performance of multi-class and multi-view unsupervised anomaly detection methods, while zero-/few-shot approaches exhibit minimal sensitivity to category variations, suggesting a fundamental shift in scalability characteristics across different learning paradigms.
\end{itemize}
\section{Related Work} \label{sec:related_work}

\subsection{IAD Datasets}

In Industrial Anomaly Detection (IAD), dataset diversity and scale are critical determinants of model effectiveness.
Current 2D IAD benchmarks, including MVTec AD~\cite{mvtec}, VisA~\cite{visa} and Real-IAD~\cite{realiad}, have demonstrated considerable success in establishing foundational evaluation frameworks. 
However, their categorical scope remains limited.
For example, Real-IAD~\cite{realiad}, the largest existing dataset, encompasses 30 categories, while MVTec AD~\cite{mvtec} and VisA~\cite{visa} contain 15 and 12 categories, respectively.
Additional datasets such as MTD~\cite{mtd}, MPDD~\cite{mpdd}, BTAD~\cite{btad}, and KolektorSDD~\cite{KolektorSDD} contribute valuable resources but \gbb{suffer from restricted scale and low categorical diversity}.

\gbb{Recent efforts have extended these benchmarks toward multi-modal or large-scale settings. For instance, MANTA~\cite{manta} introduce descriptive textual annotations for multi-modal IAD focusing on tiny objects across 38 categories. In the logistics domain, Kaputt~\cite{kaputt} presents a large-scale collection of 48,000 objects; however, its utility is constrained by the lack of pixel-level mask annotations, which are essential for precise defect localization.
Furthermore, some recent benchmarks like COCO-AD~\cite{cocoad} and ADNet~\cite{adnet} are constructed through the re-purposing or integration of existing datasets. COCO-AD~\cite{cocoad} manually redefines categories from the COCO dataset as anomalies, which fundamentally differs from the authentic ``defects" encountered in real-world industrial production. ADNet~\cite{adnet} primarily focuses on the consolidation of existing open-source benchmarks. Despite these contributions, existing datasets often exhibit limited category variability, insufficient scale, or a gap between synthetic anomalies and real-world industrial scenarios.} In contrast, our proposed Real-IAD Variety addresses these limitations by substantially enhancing category diversity, industrial scope, and defect typology, thereby advancing the field toward more comprehensive and realistic evaluation scenarios.

\subsection{Unsupervised Anomaly Detection}

Traditional unsupervised anomaly detection (UAD) algorithms \gbb{typically focuses} on single-class, single-view 2D images, \gbb{where models are trained independently} for each object category using only normal samples.
These methods are generally categorized into three paradigms:
(1) Discriminative methods (\eg, CutPaste~\cite{cutpaste}, DRAEM~\cite{draem}, and SimpleNet~\cite{simplenet}), which learn decision boundaries by distinguishing normal data from synthetically generated anomalies;
(2) Embedding-based methods (\eg, PatchCore~\cite{patchcore}, CFA~\cite{cfa}, and CSFlow~\cite{csflow}), which model the distribution of normal feature representations in a frozen or learned embedding space.
While these approaches have achieved notable success in single-class scenarios, their scalability to multi-class and cross-domain settings remains limited, motivating the development of unified detection frameworks;
(3) Reconstruction-based methods (\eg, DAE~\cite{dae}, \gbb{OCR-GAN}~\cite{ocrgan}, and RD~\cite{rd}), which operate under the assumption that anomalous regions exhibit higher reconstruction errors compared to normal patterns.

\subsection{Multi-Class Unsupervised AD}
\gbb{The Multi-class Unsupervised Anomaly Detection (MUAD) setting, pioneered by UniAD~\cite{uniad}, requires a single unified model to handle training and inference across multiple categories simultaneously.}
%To systematically evaluate the generalization capabilities of single-modality IAD models, the multi-class unsupervised anomaly detection (MUAD) setting was first introduced in UniAD~\cite{uniad}.
%This paradigm challenges models to train on all object categories within a dataset simultaneously, employing a single unified model for both training and inference.
MUAD benchmarks encompass both traditional UAD methods, such as DRAEM~\cite{draem}, SimpleNet~\cite{simplenet}, CFA~\cite{cfa}, CFLOW-AD~\cite{cflow}, and RD~\cite{rd}, explicitly designed for unified training, including UniAD~\cite{uniad}, OneNIP~\cite{gao2024onenip}, MambaAD~\cite{mambaad}, DesTSeg~\cite{destseg}, LGC~\cite{lgc} and Dinomaly~\cite{guo2025dinomaly}.
However, existing MUAD evaluations have been conducted on datasets with limited category counts (typically $\leq$ 30 classes), leaving the scalability of these methods to large-scale, highly diverse industrial scenarios largely unexplored.
Our Real-IAD Variety dataset, with its 160 categories, \gbb{provides the first rigorous testing ground for MUAD scalability}.

\subsection{Multi-View Anomaly Detection}

Beyond single-image analysis, Multi-View Anomaly Detection (MVAD) leverages complementary information from multiple viewpoints to enhance inspection accuracy and robustness.
This approach originates from the stringent precision requirements of industrial quality control, where anomalies may be occluded or imperceptible from a single perspective.
Representative methods such as MVAD~\cite{mvad} exploit multi-view consistency constraints and cross-view feature fusion to improve detection performance, which is critical for minimizing false negatives in practical deployment scenarios.
Nevertheless, existing MVAD datasets predominantly focus on limited object types and controlled imaging conditions, constraining the evaluation of cross-view generalization in diverse industrial contexts.
Real-IAD Variety addresses this gap by providing multi-view annotations across 160 categories, facilitating rigorous MVAD benchmarking.

\subsection{Zero-/Few-Shot AD Models}

Vision-Language Models (VLMs) have emerged as a promising direction for anomaly detection, enabling semantic reasoning and cross-domain generalization.
Recent advances have explored zero-shot (ZSAD) and few-shot (FSAD) anomaly detection using large-scale pre-trained models such as CLIP~\cite{clip}.
WinCLIP~\cite{winclip} introduces dual-class textual prompts and patch-wise window operation for ZSAD and FSAD.
AnomalyCLIP~\cite{iclr2024anomalyclip} learns object-agnostic prompt embeddings to capture generalizable normal and abnormal representations through an auxiliary anomaly detection dataset.
AdaCLIP~\cite{eccv24adaclip} introduces learnable hybrid prompts and regional anomaly feature extraction to improve detection precision, while VCP-CLIP~\cite{eccv24vcpclip} designs Pre-VCP and Post-VCP modules to optimize textual embeddings with visual contextual prompts, promoting cross-modal information interaction. Recently, AdaptCLIP~\cite{gao2025adaptclip} treats CLIP as a foundational service, proposing alternating and comparative learning strategies based on three lightweight adapters to support zero- and few-shot generalization across domains.

\gbb{Unlike VLM-based ZSAD and FSAD, MetaUAS~\cite{metauas} is the first to reframe anomaly detection from a change detection perspective, achieving a pure-vision one-shot anomaly detection paradigm that eliminates the reliance on specific anomaly detection datasets for pre-training.}
Despite these methodological advances, existing ZSAD and FSAD approaches have been validated primarily on small-scale datasets with limited category diversity (typically $\leq$ 30 classes), leaving their scalability and robustness under large-scale, multi-industry conditions unverified.
Our Real-IAD Variety dataset enables the first large-scale evaluation of ZSAD and FSAD methods across 160 categories, thereby facilitating systematic investigation of their generalization capabilities in realistic industrial settings.

%The MVTec Caption dataset~\cite{mvtec-caption} augments MVTec AD with descriptive textual annotations; however, it remains confined to a limited set of object types.
%Despite these contributions, existing datasets exhibit insufficient defect category variability and industrial domain coverage, thereby limiting their utility for evaluating unified and zero-/few-shot IAD paradigms.
%In contrast, our proposed Real-IAD Variety dataset addresses these limitations by substantially enhancing category diversity, industrial scope, and defect typology, thereby advancing the field toward more comprehensive and realistic evaluation scenarios.

%OneNIP~\cite{gao2024onenip} achieves robust scalability across increasing categories by utilizing a single normal image as a class-aware visual prompt to guide feature reconstruction. LGC~\cite{lgc} instead addresses the challenge of inter-class confusion by leveraging class-aware contrastive learning to enforce local and global feature consistency within the latent space. Both methods demonstrate that integrating category-level information, whether through explicit visual prompting or implicit contrastive constraints, is crucial for resolving multi-class feature reconstruction and achieving high anomaly detection performance compared to their baselines, such as UniAD~\cite{uniad} and RD~\cite{rd}, as shown in Table~\ref{tab:muad_results}.

\begin{figure*}[ht]
\centering
\includegraphics[width=1.0\linewidth]{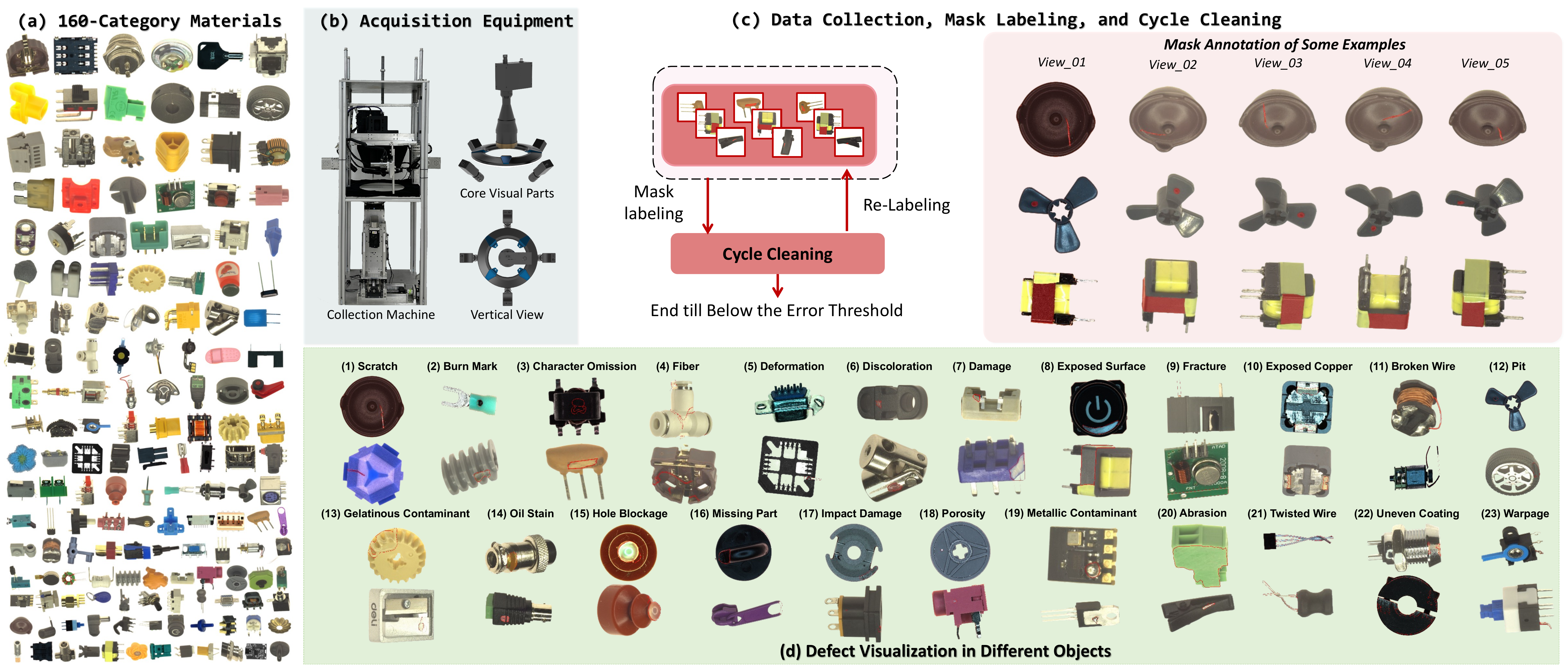}
\caption{
\textbf{Data collection and annotation pipeline for the proposed \method.}
The pipeline comprises a four-stage sequential process:
(a) \textbf{Material Preparation}: This initial phase encompasses the assembly of a diverse array of materials, spanning 160 distinct categories sourced from 28 industrial domains and encompassing 24 material compositions.
(b) \textbf{Acquisition Equipment Design}: The second phase involves the design of data capture apparatus, comprising one $\text{top-down}$ camera for overhead views and four peripheral cameras to capture lateral perspectives.
(c) \textbf{Data Collection and Annotation}: The third phase pertains to the data collection process, which includes meticulous pixel-level manual annotation, rigorous algorithmic cross-validation, and iterative refinement. \gbb{The process iterates until the variation in the model's predicted Average Precision (AP) scores falls below a predetermined threshold},
following the methodology established in Real-IAD~\cite{realiad}.
(d) \textbf{Defect Taxonomy}: The lower section illustrates distinct defect types alongside their characteristic visual representations. Zoom in for enhanced visibility of defect regions delineated in red.}
\label{fig:ppl}
\end{figure*}

\section{\method} \label{sec:method}

\subsection{Data Collection Pipeline} \label{sec:ppl}

Inspired by the methodology established in Real-IAD~\cite{realiad}, we construct a rigorous three-stage data collection pipeline to ensure dataset comprehensiveness and annotation quality, as illustrated in Fig.~\ref{fig:ppl}. The pipeline encompasses material preparation, acquisition equipment design, and data collection with iterative annotation refinement.

\noindent\textbf{Stage 1: Diverse Material Preparation.}
To construct the Real-IAD Variety dataset, a team of 12 members dedicated 11,000 working hours to material selection and procurement, simulating a wide spectrum of real-world defects based on material characteristics.
To enhance dataset representativeness, we assembled 160 object categories spanning 24 material types across 28 industrial domains.
Representative samples are depicted in Fig.~\ref{fig:ppl}a.
Given that industrial production typically achieves yield rates exceeding 99\%, naturally occurring defective parts are scarce.
Therefore, leveraging extensive production line experience, we artificially introduced four defect variations for each material category, aggregating to \gbb{27 defect types} across the entire dataset.

\begin{table}[ht]
\centering
\gbb{
\caption{Detailed Specifications of the Multi-view Acquisition System.}
\label{tab:camera_specs}
\resizebox{1.0\columnwidth}{!}{ % <--- 开始缩放：将宽度设置为页面文本宽度，高度按比例自动缩放
    \begin{tabular}{lcc}
    \toprule
    \textbf{Specification} & \textbf{Top-down Camera} & \textbf{Peripheral Cameras} \\ 
    \midrule
    Sensor Resolution & 5,328 $\times$ 3,040  & 4,096 $\times$ 3,000  \\
    Quantity & 1 & 4 \\
    Mounting Angle & 90$^\circ$ (Vertical) & 40$^\circ$--45$^\circ$ (Oblique) \\
    Lateral Accuracy & {0.01 mm/pixel} & {0.028 mm/pixel} \\
    Working Distance & 185 mm & 310 mm \\
    Field of View (FoV) & 53 $\times$ 30.4 mm & 114 $\times$ 84 mm \\
    Z-Repeatability &{$<$0.3 $\mu$m} & N/A \\
    Light Source & \multicolumn{2}{c}{RGBW Multi-spectral} \\
    \bottomrule
    \end{tabular}
} 
}% <--- 结束缩放
\end{table}

\noindent\textbf{Stage 2: Acquisition Equipment Design.}
\gbb{As illustrated in Fig.~\ref{fig:ppl}b, the acquisition apparatus is meticulously engineered to achieve high-precision surface characterization across diverse materials. The system is composed of one top-down camera, four peripheral cameras, and a shared multi-spectral light source (RGBW). The top-down camera utilizes a 5,328$\times$3,040 resolution industrial sensor positioned at a 90$^\circ$ angle, achieving a lateral accuracy of 0.01 mm/pixel and a Z-axis repeatability of $<$0.3 $\mu$m. 
To complement this, the four peripheral cameras are symmetrically arranged at 40$^\circ$-45$^\circ$ angles, each featuring a 4,096$\times$3,000 resolution sensor and a 50mm industrial lens. These units provide a broader field of view (114$\times$84 mm) with a lateral accuracy of 0.028 mm/pixel. Detailed hardware specifications are summarized in Table~\ref{tab:camera_specs}.
}

\noindent\textbf{Stage 3: Data Collection and Annotation Refinement.}
The data collection and annotation process for Real-IAD Variety adheres to the rigorous standards established by Real-IAD~\cite{realiad}.
This stage involves meticulous pixel-level manual annotation, algorithmic cross-validation, and iterative refinement.
The annotation process iterates until the model's predicted Average Precision (AP) scores exhibit negligible variation below a predetermined threshold, ensuring annotation consistency and quality.

\subsection{Comparison with Existing IAD Datasets}

\begin{table*}[!t]
    \centering
    \caption{Comparison with popular 2D IAD datasets on different attributes. \cmark: Satisfied. \xmark: Unsatisfied.}\label{tab:data_comparison}
    \resizebox{1.0\textwidth}{!}{
    
        \begin{tabular}{lcrrrcccc}
        \toprule[0.1em]
        \multirow{2}{*}{Datasets} & \multirow{2}{*}{Classes} & \multicolumn{3}{c}{Number of Images} & \multirow{2}{*}{\makecell[c]{Image \\ Resolution}} & \multirow{2}{*}{\makecell[c]{Anomaly \\ Masks}} & \multirow{2}{*}{\makecell[c]{Multiple \\ Views}}  &\multirow{2}{*}{\makecell[c]{Defect \\ Types}} \\
        \cline{3-5}
        & & Normal & Anomaly & All & & & &  \\
        \hline
        MVTec AD~\cite{mvtec} & 15 & 4,096 & 1,258 & 5,354 & 700$\sim$1,024 & \cmark  & \xmark & 20 \\
        VisA~\cite{visa} & 12 & 9,621 & 1,200 & 10,821 & 960$\sim$1,562 & \cmark & \xmark  & 13 \\
        BTAD~\cite{btad} & 3 & 2,250 & 580 & 2,830 & 600$\sim$1,600 & \cmark & \xmark  & 3 \\
        MPDD~\cite{mpdd} & 6 & 1,064 & 282 & 1,346 & 1,024$\sim$1,024 & \cmark & \xmark  & 8 \\
        MAD-Real~\cite{pad} & 10 & 540 & 221 & 761 & 3,472$\sim$3,472 & \cmark & \cmark  & 2\\
        MAD-Sim~\cite{pad} & 20 & 4,838 & 4,951 & 9,789 & 800$\sim$800 & \cmark & \cmark  & 3\\
        Real-IAD~\cite{realiad} & 30 & 99,721 & 51,329 & 151,050 & 2,000$\sim$5,000 & \cmark & \cmark  & 8 \\
        \gbb{MANTA~\cite{manta}} & 38  &{652,455} &34,235 &{686,690} & 162$\sim$3,365 & \cmark & \cmark  & {30}  \\
        \gbb{Kaputt~\cite{kaputt}} &{48,376} & 70,951 & 29,316 & 100,267 & 2,048$\sim$2,048 & \xmark & \xmark  & 7 \\
        \hline
        \rowcolor{blue_tab}
        Real-IAD Variety & 160
        &39,905	&{159,045}	&{198,950}  &{260$\sim$5,328} & \cmark & \cmark  & {27} \\
        \hline
        \end{tabular}
        }
\end{table*}

Table~\ref{tab:data_comparison} presents a comprehensive comparison between Real-IAD Variety and mainstream IAD datasets.
Real-IAD Variety is the first IAD dataset to scale object categories to the hundreds, reaching 160 classes, thereby opening new avenues for large-scale multi-class anomaly detection research analogous to the impact of ImageNet~\cite{imagenet} in object recognition.
Compared to existing benchmarks, Real-IAD Variety offers several distinctive advantages:
(1) Unprecedented scale: With \gbb{198,950} images including \gbb{159,045} anomalous images, it provides substantially larger training and evaluation data compared to prior datasets;
(2) Comprehensive annotations: Pixel-level defect masks with rigorous quality control ensure high annotation fidelity;
(3) Multi-view coverage: Multiple viewpoints per sample enable robust evaluation of view-invariant detection methods. \gbb{Note that pixel-level masks are precisely annotated only in the specific views where defects are visible, as shown in Fig.~\ref{fig:typical_samples}};
(4) Diverse defect taxonomy: \gbb{27 defect types} represent the most comprehensive defect categorization among existing IAD datasets;
(5) Extended resolution range: Image resolutions spanning, \ie, 260$\sim$5,328 (Fig.~\ref{fig:imageresolution}), accommodate industrial products of varying scales, with over 90\% of images exceeding 2,000 pixels in resolution.
\gbb{The lower bound of the resolution range (260 pixels) represents a small fraction of images where objects were tightly cropped to ensure focus on the primary region of interest.}
This wide resolution range stems from the diverse scales of industrial products, which are cropped such that the primary object occupies the majority (>90\%) of the image area.

\begin{figure*}[ht]
\centering
\includegraphics[width=1.0\linewidth]{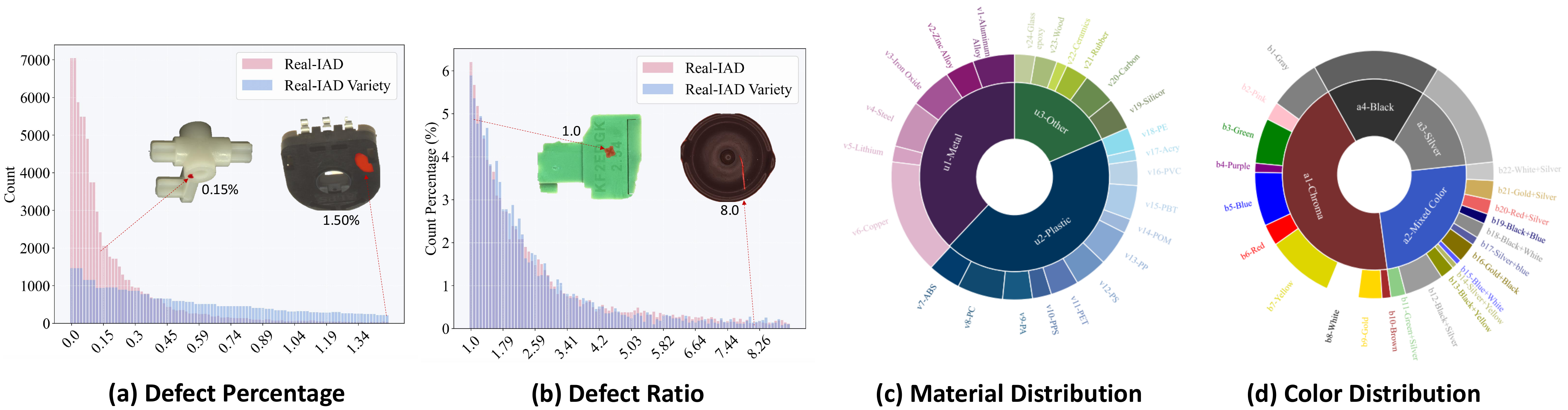}
\caption{
\textbf{Statistical characteristics of Real-IAD Variety across multiple dimensions.}
(a) \textbf{Anomalous region proportion}: Real-IAD Variety exhibits a broader and more balanced distribution of anomalous region proportions relative to total image area compared to Real-IAD~\cite{realiad}, substantially increasing dataset complexity.
(b) \textbf{Defect aspect ratio}: Real-IAD Variety provides diverse aspect ratios for minimum bounding rectangles of defects, comparable to Real-IAD, introducing additional diversity and detection challenges. Representative samples are shown for intuitive visualization.
(c) \textbf{Material distribution}: Real-IAD Variety encompasses 24 material types for practical applications, imposing higher requirements on method robustness.
(d) \textbf{Color distribution}: Real-IAD Variety captures a wide color spectrum, which is essential for color-based anomaly detection research.
}
\label{fig:statistic}
\end{figure*}

\begin{figure*}[ht]
\centering
\includegraphics[width=1.0\linewidth]{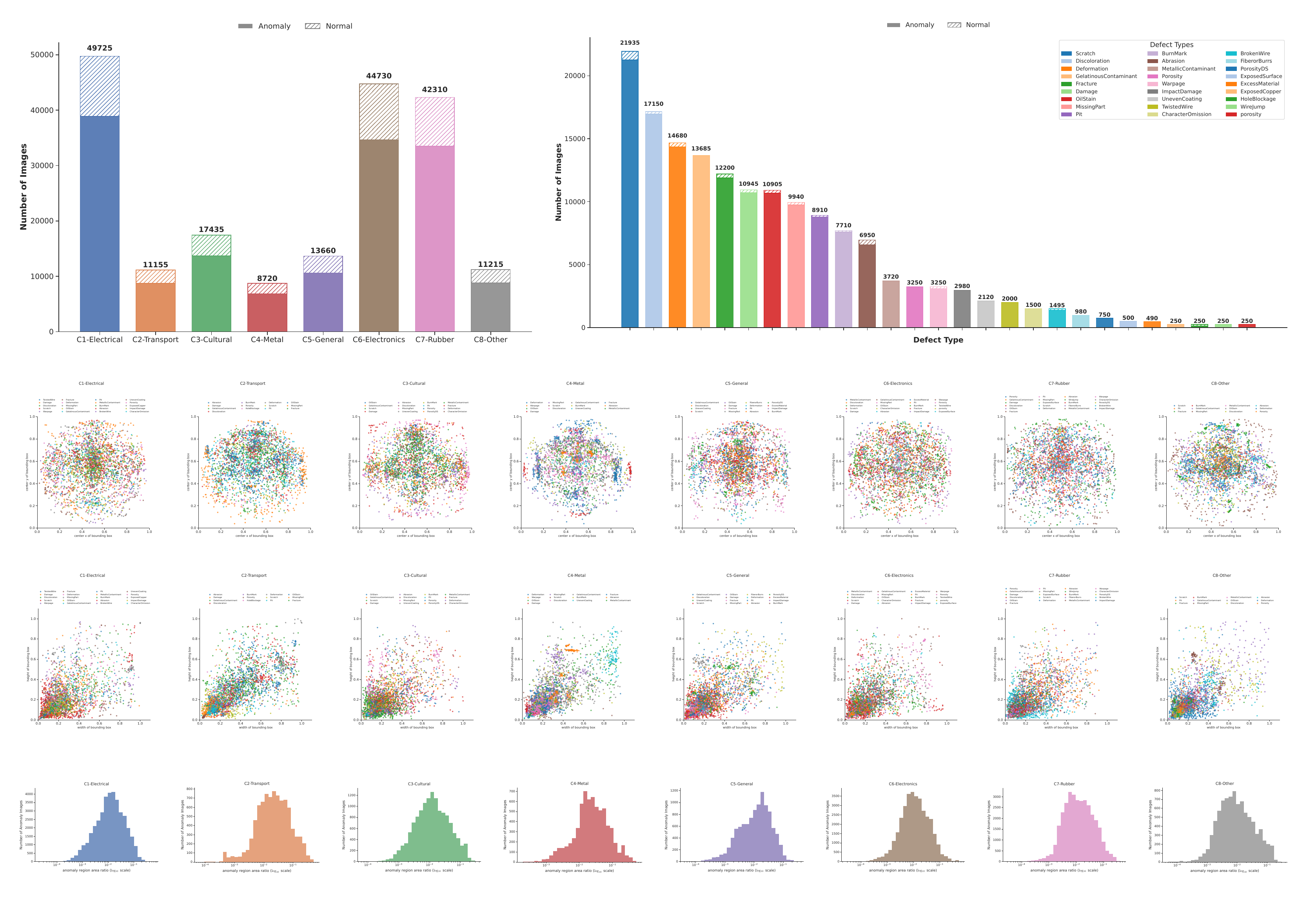}
\caption{The distribution of normal and abnormal images (first row), defect spatial locations (second row),  defects scales (third row) and defect area ratios (last row) in our Real-IAD Variety dataset.
}
\label{fig:distributionstatistic}
\end{figure*}

\subsection{Dataset Characteristics and Statistical Analysis} \label{sec:statistic}

\noindent\textbf{Statistical Distribution Analysis.}
\gbb{To provide a comprehensive understanding of Real-IAD Variety, we perform a multi-dimensional statistical analysis as illustrated in Fig.~\ref{fig:statistic} and Fig.~\ref{fig:distributionstatistic}:}
\textit{(a) Material Diversity:} Real-IAD Variety encompasses 24 material types (Fig.~\ref{fig:statistic}c) across 28 industrial domains (Fig.~\ref{fig:teaser}), providing comprehensive coverage from a practical application perspective.
\textit{(b) Color Spectrum}: The dataset captures a wide color distribution (Fig.~\ref{fig:statistic}d), which is critical for evaluating color-based anomaly detection approaches.
\textit{(c) Anomalous Region Proportion}: Compared to Real-IAD~\cite{realiad}, Real-IAD Variety exhibits a broader and more uniform distribution of anomaly region proportions (Fig.~\ref{fig:statistic}a), encompassing defects of varying sizes and thereby increasing detection difficulty.
\textit{(d) Defect Aspect Ratio}: The defect aspect ratio distribution in Real-IAD Variety is comparably diverse to Real-IAD (Fig.~\ref{fig:statistic}b), presenting substantial challenges for models to discriminate defects of various geometric configurations.
\gbb{\textit{(e) Distribution of Normal and Anomalous Images}: Real-IAD Variety contains a substantially more anomalous images than normal images, addressing the ``data scarcity" bottleneck in real-world industrial environments (first row of Fig.~\ref{fig:distributionstatistic}).
\textit{(f) Defects Spatial and Scale Distribution:} 
The spatial distribution of defect centers (second row in Fig.~\ref{fig:distributionstatistic}) reveals that anomalies occur across almost all possible image coordinates, demonstrating high spatial diversity. The defect bounding boxes relative to the original resolution shows a high density near the origin (third row in Fig.~\ref{fig:distributionstatistic}), indicating a predominance of challenging small-scale defects. \textit{(g) Anomalous Area Ratio Distribution:}
The distribution of anomalous regions (fourth row in Fig.~\ref{fig:distributionstatistic}) approximately follows a normal distribution on a logarithmic scale.}

\noindent\textbf{Representative Defect Visualization.}
Figs.~\ref{fig:typical_samples} and~\ref{fig:ppl}d presents qualitative visualizations of 27 distinct defect types across various object categories.
Notably, identical defect types manifest in diverse visual forms across different materials, reflecting the complexity of real-world industrial scenarios.
These defect types, derived from authentic production line observations, enhance the practical applicability of Real-IAD Variety.

\noindent\textbf{Advantages and Contributions.}
Derived from real production environments, Real-IAD Variety offers several distinctive advantages:
(1) Material diversity: 24 material types representing the most comprehensive material coverage among existing IAD datasets;
(2) Defect types: 27 defect types providing extensive defect variation;
(3) Resolution range: Extended image resolutions (260$\sim$5,328 pixels) accommodating diverse industrial product scales;
(4) Multi-view annotations: Multiple viewpoints per sample enabling view-invariant evaluation;
(5) Annotation quality: Highly accurate pixel-level mask annotations with rigorous quality control;
Real-IAD Variety represents the first IAD dataset to systematically consider comprehensive coverage of material types and color variations, thereby robustly supporting the development of large-scale IAD models, industrial foundation models, vision-language IAD frameworks, and high-resolution detection paradigms.
\section{Experiments} \label{sec:exp}

\subsection{Dataset Protocol and Evaluation Metrics} \label{sec:metrics}

\noindent{\textbf{Training and Testing Protocol}}.
The Real-IAD Variety dataset is partitioned into training and testing subsets, comprising 3,991 normal samples (19,955 images) for training and 35,799 samples (3,990 normal and 31,809 anomalous, totaling 178,995 images) for testing.
Notably, the testing subset exhibits a nearly balanced distribution between normal and anomalous images.
To systematically evaluate the impact of category count on model performance, we partition Real-IAD Variety into three subsets: S1, S2, and S3, containing 30, 60, and 100 categories, respectively.
The category selection employs randomization to mitigate potential biases in color and material representation.

\begin{table*}[t]
\centering
\gbb{
\caption{Quantitative MUAD performance comparisons on \method~dataset. Dis., Emb., and Rec. represent discrimination-based, embedding-based, and reconstruction-based methods, respectively. The best results are indicated in \textbf{bold}.}
\label{tab:muad_results}
\resizebox{1.8\columnwidth}{!}{%
\begin{tabular}{lllc@{\hspace{6pt}}c@{\hspace{6pt}}c@{\hspace{6pt}}c@{\hspace{6pt}}c@{\hspace{6pt}}c@{\hspace{6pt}}c}
\toprule
\textbf{Paradigms} & \textbf{Methods} & \textbf{Backbone} & \textbf{I-ROC} & \textbf{I-PR} & \textbf{I-F1$_{max}$} & \textbf{P-ROC} & \textbf{P-PR} & \textbf{P-F1$_{max}$} & \textbf{P-PRO} \\ 
\midrule
\multirow{2}{*}{Dis.}  & DRAEM~\cite{draem} & scratch & 54.9 & 89.3 & 93.1 & 53.7 & 9.4 & 8.8 & 21.6 \\
& SimpleNet~\cite{simplenet} & WRes-50 & 53.7 & 88.5 & 93.0 & 62.6 & 4.8 & 8.8 & 29.5 \\
\midrule
\multirow{2}{*}{Emb.}  & CFA~\cite{cfa} & WRes-50 & 52.9 & 88.5 & 93.0 & 52.9 & 2.7 & 5.6 & 14.1 \\
 & CFLOW~\cite{cflow} & WRes-50 & 63.8 & 91.7 & 93.1 & 85.2 & 14.8 & 20.0 & 59.5 \\
 \midrule
\multirow{9}{*}{Rec.} & UniAD~\cite{uniad} & Eff-B4 & 67.5 & 92.4 & 93.3 & 87.1 & 18.0 & 24.0 & 62.8 \\
 & OneNIP~\cite{gao2024onenip} & Eff-B4 & 73.3 & 94.2 & 93.5 & 88.3 & 24.0 & 29.4 & 64.5 \\
 & RD~\cite{rd} & WRes-50 & 72.8 & 93.9 & 93.6 & 88.9 & 22.3 & 28.1 & 68.9 \\
 & DeSTSeg~\cite{destseg} & Res-18 & 75.0 & 95.0 & 93.2 & 65.6 & 37.2 & 31.2 & 37.7 \\
 & MambaAD~\cite{mambaad} & Res-34 & 81.7 & 96.2 & 93.9 & 91.2 & 33.1 & 37.7 & 73.5 \\
 & LGC~\cite{lgc} & WRes-50 & 81.0 & 96.0 & 94.0 & 91.7 & 31.9 & 37.0 & 74.6 \\
 & Dinomaly~\cite{guo2025dinomaly} & ViT-B-14 & 85.4 & 97.2 & \textbf{94.5} & 91.5 & 42.8 & 45.8 & 75.6 \\
 & Dinomaly+ (Ours) & ViT-B-14 & \textbf{87.1} & \textbf{97.3} & {94.0} & \textbf{91.9} & \textbf{49.7} & \textbf{49.2} & \textbf{76.4} \\ 
\bottomrule
\end{tabular}%
}
}
\end{table*}

\noindent{\textbf{Evaluation Metrics}}.
Following established anomaly detection protocols, \gbb{we employ seven standard metrics to quantify performance across both image-level anomaly classification and pixel-level anomaly localization tasks. These seven metrics include: Image-level Area Under the Receiver Operating Characteristic curve (I-ROC), Area Under the Precision-Recall curve (I-PR), and Maximum F1-score (I-F1\({}_{max}\)); alongside their pixel-level counterparts (P-ROC, P-PR, P-F1\({}_{max}\)) and Pixel-level Per-Region Overlap (P-PRO) for fine-grained defect localization assessment.}

\subsection{Benchmark on Multi-Class Unsupervised AD}
\noindent\textbf{Experimental Setting}.
The Multi-Class Unsupervised Anomaly Detection (MUAD) paradigm, first introduced in UniAD~\cite{uniad}, trains a unified model on all object categories within a dataset simultaneously, employing the same model for both training and inference.
This approach eliminates the need for class-specific models, thereby reducing storage costs and potentially enabling the learning of generalizable features across multiple categories.
This setting holds significant practical value for real-world industrial deployment.

Unless otherwise specified, MUAD models are \gbb{trained at a resolution of 256$\times$256 for 100 epochs}. For Dinomaly \cite{guo2025dinomaly}, we follow the original configuration, \gbb{utilizing a 392$\times$392 crop} from 448$\times$448 input resolution for 50,000 steps (\eg, approximately 160 epochs for Real-IAD Variety).
\gbb{Additionally, we introduce Dinomaly+, an enhanced version that extends the baseline Dinomaly \cite{guo2025dinomaly} by integrating a refined segmentation head~\cite{gao2024onenip}.} This model is fine-tuned for 4 epochs on pseudo-anomalous data to facilitate high-precision, coarse-to-fine anomaly localization.

\noindent\textbf{Competitive Methods}.
We evaluate state-of-the-art unsupervised anomaly detection methods to comprehensively assess their scalability to large-scale datasets.
Specifically, we benchmark discrimination-based methods (DRAEM~\cite{draem}, SimpleNet~\cite{simplenet}), embedding-based methods (CFA~\cite{cfa}, CFLOW-AD~\cite{cflow}), and reconstruction-based methods (UniAD~\cite{uniad}, OneNIP~\cite{gao2024onenip}, RD~\cite{rd}, DesTSeg~\cite{destseg}, MambaAD~\cite{mambaad}, LGC~\cite{lgc}, Dinomaly~\cite{guo2025dinomaly}, and Dinomaly+) on Real-IAD Variety.
Considering computational resource constraints, we focus on representative and strong baselines, SimpleNet~\cite{simplenet}, CFLOW-AD~\cite{cflow},  UniAD~\cite{uniad}, OneNIP~\cite{gao2024onenip}, MambaAD~\cite{mambaad}, LGC~\cite{lgc},  Dinomaly~\cite{guo2025dinomaly}, and Dinomaly+ for comprehensive MUAD evaluation.
Notably, PatchCore~\cite{patchcore} is excluded from this comparison due to its memory bank mechanism, which incurs prohibitive memory consumption as category count increases, leading to potential out-of-memory issues on large-scale datasets like Real-IAD Variety.

\begin{table}[h]
\centering
%\begingroup 
%\color{red}
\caption{Comprehensive comparison of MUAD methods on Real-IAD and Real-IAD Variety datasets. $\Delta$ represents the absolute difference between the maximum and minimum values across Real-IAD Variety scales (S1, S2, S3, and Full) for each method, highlighting performance sensitivity to category expansion, and the same below.}
\label{tab:muad_multicatergories}
\setlength{\tabcolsep}{3.5pt}
\resizebox{1.0\columnwidth}{!}{%
\begin{tabular}{lccccccccc}
\toprule
\textbf{Methods} & \textbf{Datasets} & \textbf{Classes} & \textbf{I-ROC} & \textbf{I-PR} & \textbf{I-F1$_{max}$} & \textbf{P-ROC} & \textbf{P-PR} & \textbf{P-F1$_{max}$} & \textbf{P-PRO} \\ 
\midrule
\multirow{6}{*}{SimpleNet~\cite{simplenet}} 
& Real-IAD & 30 & 54.9 & 50.6 & 61.5 & 76.1 & 1.9 & 4.9 & 42.4 \\
& Real-IAD Variety S1 & 30 & 75.4 & 95.2 & 93.7 & 83.4 & 22.1 & 26.6 & 52.7 \\
& Real-IAD Variety S2 & 60 & 67.9 & 93.2 & 93.2 & 80.4 & 14.4 & 19.2 & 46.0 \\
& Real-IAD Variety S3 & 100 & 61.0 & 91.0 & 93.1 & 78.0 & 9.2 & 13.8 & 41.8 \\
& Real-IAD Variety & 160 & 53.7 & 88.5 & 93.0 & 62.6 & 4.8 & 8.8 & 29.5 \\
\rowcolor[gray]{0.95} \multicolumn{3}{c}{$\Delta \downarrow$} & 21.7 & 6.7 & 0.7 & 20.8 & 17.3 & 17.8 & 23.2 \\ \midrule
\multirow{6}{*}{CFLOW~\cite{cflow}} 
& Real-IAD & 30 & 77.0 & 75.8 & 69.9 & 94.8 & 17.6 & 21.7 & 80.4 \\
& Real-IAD Variety S1 & 30 & 77.9 & 95.4 & 93.8 & 87.5 & 27.6 & 26.6 & 63.2 \\
& Real-IAD Variety S2 & 60 & 72.5 & 94.1 & 93.3 & 86.7 & 23.6 & 24.5 & 61.3 \\
& Real-IAD Variety S3 & 100 & 72.5 & 94.1 & 93.3 & 86.7 & 23.6 & 24.5 & 61.3 \\
& Real-IAD Variety & 160 & 63.8 & 91.7 & 93.1 & 85.2 & 14.8 & 20.0 & 59.5 \\
\rowcolor[gray]{0.95} \multicolumn{3}{c}{$\Delta \downarrow$} & 14.1 & 3.7 & 0.7 & 2.3 & 12.8 & 6.6 & 3.7 \\ \midrule

\multirow{6}{*}{UniAD~\cite{uniad}} 
& Real-IAD & 30 & 83.1 & 81.2 & 74.5 & 97.4 & 23.3 & 30.9 & 87.1 \\
& Real-IAD Variety S1 & 30 & 78.5 & 95.4 & 94.2 & 88.9 & 29.5 & 34.1 & 67.1 \\
& Real-IAD Variety S2 & 60 & 71.6 & 93.5 & 93.7 & 87.3 & 21.7 & 27.6 & 64.8 \\
& Real-IAD Variety S3 & 100 & 70.2 & 93.1 & 93.5 & 87.6 & 19.0 & 25.0 & 64.3 \\
& Real-IAD Variety & 160 & 67.5 & 92.4 & 93.3 & 87.1 & 18.0 & 24.0 & 62.8 \\
\rowcolor[gray]{0.95} \multicolumn{3}{c}{$\Delta \downarrow$} & 11.0 & 3.0 & 0.9 & 1.8 & 11.5 & 10.1 & 4.3 \\ \midrule

\multirow{6}{*}{OneNIP~\cite{gao2024onenip}} 
& Real-IAD & 30 & 86.3 & 84.9 & 77.2 & 98.2 & 37.0 & 41.5 & 88.6 \\
& Real-IAD Variety S1 & 30 & 82.8 & 96.7 & 94.4 & 90.1 & 36.7 & 39.1 & 69.0 \\
& Real-IAD Variety S2 & 60 & 78.7 & 95.6 & 94.0 & 89.2 & 30.7 & 34.9 & 67.2 \\
& Real-IAD Variety S3 & 100 & 76.5 & 94.9 & 93.7 & 89.2 & 27.0 & 32.2 & 66.5 \\
& Real-IAD Variety & 160 & 73.3 & 94.2 & 93.5 & 88.3 & 24.0 & 29.4 & 64.5 \\
\rowcolor[gray]{0.95} \multicolumn{3}{c}{$\Delta \downarrow$} & 9.5 & 2.5 & 0.9 & 1.8 & 12.7 & 9.7 & 4.5 \\ \midrule

\multirow{6}{*}{MambaAD~\cite{mambaad}} 
& Real-IAD & 30 & 87.0 & 85.3 & 77.6 & 98.6 & 32.4 & 38.1 & 91.2 \\
& Real-IAD Variety S1 & 30 & 87.4 & 97.5 & 95.0 & 91.5 & 40.3 & 43.2 & 75.0 \\
& Real-IAD Variety S2 & 60 & 76.2 & 94.9 & 93.8 & 89.0 & 25.9 & 31.5 & 68.1 \\
& Real-IAD Variety S3 & 100 & 72.5 & 93.7 & 93.6 & 88.5 & 21.5 & 27.5 & 66.5 \\
& Real-IAD Variety & 160 & 81.7 & 96.2 & 93.9 & 91.2 & 33.1 & 37.7 & 73.5 \\
\rowcolor[gray]{0.95} \multicolumn{3}{c}{$\Delta \downarrow$} & 14.9 & 3.8 & 1.4 & 3.0 & 18.8 & 15.7 & 8.5 \\ \midrule

\multirow{6}{*}{LGC~\cite{lgc}} 
& Real-IAD & 30 & 85.9 & 83.0 & 77.1 & 98.1 & 25.3 & 33.3 & 90.7 \\
& Real-IAD Variety S1 & 30 & 84.1 & 96.7 & 94.8 & 91.0 & 36.1 & 39.9 & 73.7 \\
& Real-IAD Variety S2 & 60 & 82.7 & 96.4 & 94.1 & 91.9 & 34.5 & 39.9 & 76.6 \\
& Real-IAD Variety S3 & 100 & 82.4 & 96.2 & 94.3 & 91.8 & 32.8 & 37.9 & 75.0 \\
& Real-IAD Variety & 160 & 81.0 & 96.0 & 94.0 & 91.7 & 31.9 & 37.0 & 74.6 \\
\rowcolor[gray]{0.95} \multicolumn{3}{c}{$\Delta \downarrow$} & 3.1 & 0.7 & 0.8 & 0.9 & 4.2 & 2.9 & 2.9 \\ \midrule

\multirow{6}{*}{Dinomaly~\cite{guo2025dinomaly}} 
& Real-IAD & 30 & 89.3 & 86.5 & 80.1 & 98.9 & 42.8 & 47.1 & 93.9 \\
& Real-IAD Variety S1 & 30 & 91.4 & 98.4 & 95.8 & 92.7 & 56.9 & 56.1 & 79.8 \\
& Real-IAD Variety S2 & 60 & 89.7 & 98.1 & 95.2 & 92.6 & 53.1 & 53.7 & 79.7 \\
& Real-IAD Variety S3 & 100 & 87.9 & 97.7 & 94.9 & 92.3 & 48.3 & 50.1 & 78.2 \\
& Real-IAD Variety & 160 & 85.4 & 97.2 & 94.5 & 91.5 & 42.8 & 45.8 & 75.6 \\
\rowcolor[gray]{0.95} \multicolumn{3}{c}{$\Delta \downarrow$} & 6.0 & 1.2 & 1.4 & 1.2 & 14.1 & 10.3 & 4.2 \\ \midrule

\multirow{6}{*}{Dinomaly+} 
& Real-IAD & 30 & 89.5 & 86.3 & 80.7 & 98.8 & 45.6 & 48.6 & 93.1 \\
& Real-IAD Variety S1 & 30 & 92.4 & 98.6 & 96.0 & 92.7 & 59.2 & 56.1 & 79.3 \\
& Real-IAD Variety S2 & 60 & 90.8 & 98.3 & 95.3 & 92.9 & 56.3 & 54.4 & 79.6 \\
& Real-IAD Variety S3 & 100 & 89.4 & 97.9 & 94.7 & 92.4 & 53.6 & 52.5 & 78.4 \\
& Real-IAD Variety & 160 & 87.1 & 97.3 & 94.0 & 91.9 & 49.7 & 49.2 & 76.4 \\
\rowcolor[gray]{0.95} \multicolumn{3}{c}{$\Delta \downarrow$} & 5.3 & 1.3 & 2.0 & 6.9 & 9.5 & 6.9 & 3.2 \\ 
\bottomrule
\end{tabular}
}
%\endgroup %
\end{table}

\noindent\textbf{Results and Analysis under MUAD}.
Table~\ref{tab:muad_results} presents quantitative results of different methods on Real-IAD Variety.
Several critical observations emerge that differ from findings on small-scale datasets.
(1) All evaluated methods experience substantial and non-saturating performance degradation on Real-IAD Variety, indicating the dataset's significant challenge for future research and confirming successful avoidance of the metric saturation phenomenon frequently observed in smaller-scale IAD benchmarks;
(2) Discrimination-based approaches (DRAEM~\cite{draem}, SimpleNet~\cite{simplenet}) and the embedding-based method CFA~\cite{cfa} exhibit near-failure performance, with I-ROC scores approaching the 50\% random guessing threshold. This demonstrates their severe lack of generalization capability when confronted with large-scale, high-variance normal data distributions;
(3) Reconstruction-based methods maintain relatively satisfactory performance despite the challenging conditions. Notably, Dinomaly+ (building upon Dinomaly~\cite{guo2025dinomaly} and OneNIP~\cite{gao2024onenip}) achieves the best overall results across most metrics \gbb{(\eg, 87.1\% I-ROC and 49.7\% P-PR)}, underscoring its superior generalization and robustness across large-scale multi-class scenarios. In contrast, UniAD~\cite{uniad}, an early MUAD method, exhibits one of the most severe performance drops among reconstruction-based approaches.

\noindent\textbf{Performance Trends with Increasing Categories}.
To evaluate performance trends as category count increases, we selected eight representative methods across different paradigms.
Table~\ref{tab:muad_multicatergories} presents results on Real-IAD and Real-IAD Variety with incrementally increasing categories: S1 (30 classes), S2 (60 classes), S3 (100 classes), and the full 160 classes (visualized in Fig.~\ref{fig:incre}).

Several consistent conclusions emerge under the MUAD setting.
(1) \gbb{Real-IAD Variety poses a more significant challenge than the original Real-IAD. Even at the same category scale (30 classes), the increased intra-class variance and higher resolution complexity in Real-IAD Variety lead to consistent performance drops across most baselines.} 
(2) A fundamental scalability challenge of MUAD is revealed: consistent performance degradation occurs across all evaluated methods as category count increases. Crucially, the magnitude of degradation is method-dependent, highlighting that different architectures possess varying levels of category generalization capability and robustness to scale-up.
(3) Methods leveraging powerful backbones (\eg, DINOv2-R~\cite{dinov2r}) and advanced architectures (\eg, Transformer decoders), such as Dinomaly and Dinomaly+, consistently achieve substantial performance superiority across all datasets. This advantage is especially prominent on high-scale benchmarks, underscoring their effectiveness over normalizing flow-based (\eg, CFLOW-AD) or simpler backbone-based models (\eg, SimpleNet, UniAD, MambaAD).

\gbb{The performance degradation of MUAD methods when scaling to a large number of categories primarily stems from two factors.
First, as the number of categories increases, models often suffer from a capacity-diversity conflict, converging toward a ``generic'' feature space that loses category-specific nuances essential for spotting subtle anomalies. To address this, recent methods such as OneNIP~\cite{gao2024onenip} and LGC~\cite{lgc} introduce category-level information via explicit visual prompts or class-aware contrastive learning to resolve feature reconstruction and preserve class-specific details. Second, reconstruction-based multi-class models are susceptible to identity mapping. In these cases, the model's excessive generalization capability allows it to reconstruct the anomalies themselves rather than restoring them to a normal state, thereby failing to produce a significant residual for detection.
The state-of-the-art Dinomaly~\cite{guo2025dinomaly} partially mitigates this by leveraging a more powerful vision encoder (\eg, DINOv2-R) and simple dropout to achieve more robust performance.}

\begin{table}[t]
\centering
%\begingroup % 
%\color{red}
\caption{Quantitative comparison of ZSAD and FSAD methods on Real-IAD and Real-IAD Variety datasets. Results for FSAD methods represent the average and standard deviation over three runs with random seeds.}
\label{tab:zsad_multicatergories}
\setlength{\tabcolsep}{2.5pt} % 
\resizebox{1.0\columnwidth}{!}{%
\begin{tabular}{llccccccccc}
\toprule
\textbf{Methods} & \textbf{Datasets} & \textbf{Classes} & \textbf{shot} & \textbf{I-ROC} & \textbf{I-PR} & \textbf{I-F1$_{max}$} & \textbf{P-ROC} & \textbf{P-PR} & \textbf{P-F1$_{max}$} & \textbf{P-PRO} \\ 
\midrule
\multirow{6}{*}{AnomalyCLIP~\cite{iclr2024anomalyclip}} 
& Real-IAD & 30 & 0 & 69.1 & 63.5 & 65.4 & 95.2 & 26.8 & 34.7 & 84.9 \\
& Real-IAD Variety S1 & 30 & 0 & 68.6 & 88.9 & 88.6 & 88.7 & 36.5 & 40.3 & 71.7 \\
& Real-IAD Variety S2 & 60 & 0 & 68.8 & 88.6 & 88.3 & 88.5 & 36.1 & 40.0 & 72.4 \\
& Real-IAD Variety S3 & 100 & 0 & 69.5 & 88.9 & 88.4 & 89.5 & 36.2 & 40.2 & 74.2 \\
& Real-IAD Variety & 160 & 0 & 69.3 & 88.9 & 88.3 & 89.3 & 35.3 & 39.6 & 74.6 \\
\rowcolor[gray]{0.95} \multicolumn{4}{c}{$\Delta \downarrow$} & 0.9 & 0.3 & 0.3 & 1.0 & 1.2 & 0.7 & 2.9 \\ \midrule

\multirow{6}{*}{AdaCLIP~\cite{eccv24adaclip}} 
& Real-IAD & 30 & 0 & 71.1 & 67.2 & 67.0 & 95.6 & 33.6 & 38.1 & 90.2 \\
& Real-IAD Variety S1 & 30 & 0 & 69.7 & 89.3 & 88.7 & 87.9 & 30.8 & 35.4 & 70.0 \\
& Real-IAD Variety S2 & 60 & 0 & 70.1 & 88.9 & 88.7 & 88.0 & 32.0 & 36.6 & 71.3 \\
& Real-IAD Variety S3 & 100 & 0 & 71.3 & 89.4 & 88.6 & 88.9 & 32.8 & 37.0 & 72.7 \\
& Real-IAD Variety & 160 & 0 & 71.1 & 89.3 & 88.6 & 89.1 & 32.6 & 36.8 & 73.1 \\
\rowcolor[gray]{0.95} \multicolumn{4}{c}{$\Delta \downarrow$} & 1.6 & 0.5 & 0.1 & 1.2 & 2.0 & 1.6 & 3.1 \\ \midrule

\multirow{6}{*}{VCPCLIP~\cite{eccv24vcpclip}} 
& Real-IAD & 30 & 0 & 72.5 & 70.9 & 68.2 & 95.9 & 28.8 & 34.5 & 83.7 \\
& Real-IAD Variety S1 & 30 & 0 & 71.4 & 90.0 & 88.7 & 87.3 & 34.9 & 38.1 & 67.2 \\
& Real-IAD Variety S2 & 60 & 0 & 73.0 & 90.4 & 88.7 & 87.9 & 37.5 & 40.3 & 69.5 \\
& Real-IAD Variety S3 & 100 & 0 & 72.9 & 90.5 & 88.6 & 89.1 & 37.6 & 40.4 & 71.1 \\
& Real-IAD Variety & 160 & 0 & 72.6 & 90.2 & 88.6 & 89.3 & 36.9 & 39.9 & 70.9 \\
\rowcolor[gray]{0.95} \multicolumn{4}{c}{$\Delta \downarrow$} & 1.6 & 0.5 & 0.1 & 2.0 & 2.7 & 2.3 & 3.9 \\ \midrule

\multirow{6}{*}{AdaptCLIP~\cite{gao2025adaptclip}} 
& Real-IAD & 30 & 0 & 74.2 & 70.8 & 68.9 & 94.9 & 27.9 & 35.4 & 83.1 \\
& Real-IAD Variety S1 & 30 & 0 & 73.2 & 94.7 & 93.7 & 88.0 & 36.5 & 40.0 & 68.3 \\
& Real-IAD Variety S2 & 60 & 0 & 72.4 & 94.2 & 93.5 & 88.1 & 36.3 & 39.9 & 69.2 \\
& Real-IAD Variety S3 & 100 & 0 & 72.6 & 94.2 & 93.4 & 89.0 & 36.6 & 40.1 & 71.1 \\
& Real-IAD Variety & 160 & 0 & 72.9 & 94.3 & 93.4 & 89.1 & 36.1 & 39.9 & 70.8 \\
\rowcolor[gray]{0.95} \multicolumn{4}{c}{$\Delta \downarrow$} & 0.4 & 0.5 & 0.3 & 1.1 & 0.5 & 0.2 & 2.8 \\ \midrule

\multirow{6}{*}{MetaUAS~\cite{metauas}} 
& Real-IAD & 30 & 1 & 80.0$\pm$0.4 & 77.9$\pm$0.4 & 72.4$\pm$0.4 & 95.6$\pm$0.2 & 36.6$\pm$1.1 & 39.7$\pm$1.0 & 83.5$\pm$0.7 \\
& Real-IAD Variety S1 & 30 & 1 & 80.0$\pm$0.2 & 96.1$\pm$0.1 & 94.3$\pm$0.0 & 90.5$\pm$0.2 & 49.0$\pm$0.9 & 48.3$\pm$0.7 & 73.3$\pm$0.1 \\
& Real-IAD Variety S2 & 60 & 1 & 80.3$\pm$0.3 & 96.0$\pm$0.1 & 94.1$\pm$0.1 & 91.4$\pm$0.1 & 46.7$\pm$0.2 & 47.1$\pm$0.2 & 74.3$\pm$0.0 \\
& Real-IAD Variety S3 & 100 & 1 & 81.7$\pm$0.3 & 96.1$\pm$0.0 & 94.2$\pm$0.1 & 91.9$\pm$0.0 & 48.1$\pm$0.5 & 48.2$\pm$0.4 & 76.0$\pm$0.2 \\
& Real-IAD Variety & 160 & 1 & 81.9$\pm$0.1 & 96.3$\pm$0.1 & 94.1$\pm$0.0 & 92.0$\pm$0.1 & 48.2$\pm$0.4 & 48.3$\pm$0.3 & 76.5$\pm$0.1 \\
\rowcolor[gray]{0.95} \multicolumn{4}{c}{$\Delta \downarrow$} & 1.9 & 0.3 & 0.2 & 1.5 & 2.3 & 1.2 & 3.2 \\ \midrule

\multirow{6}{*}{AdaptCLIP~\cite{gao2025adaptclip}} 
& Real-IAD & 30 & 1 & 81.7$\pm$0.3 & 80.2$\pm$0.2 & 73.4$\pm$0.1 & 97.1$\pm$0.1 & 36.2$\pm$0.3 & 42.2$\pm$0.3 & 88.1$\pm$0.2 \\
& Real-IAD Variety S1 & 30 & 1 & 83.3$\pm$0.3 & 96.9$\pm$0.0 & 94.4$\pm$0.0 & 91.1$\pm$0.1 & 49.4$\pm$0.1 & 50.0$\pm$0.1 & 74.2$\pm$0.1 \\
& Real-IAD Variety S2 & 60 & 1 & 82.9$\pm$0.6 & 96.6$\pm$0.1 & 94.2$\pm$0.1 & 91.4$\pm$0.0 & 48.2$\pm$0.6 & 49.0$\pm$0.5 & 75.6$\pm$0.1 \\
& Real-IAD Variety S3 & 100 & 1 & 83.8$\pm$0.3 & 96.7$\pm$0.1 & 94.2$\pm$0.0 & 92.2$\pm$0.1 & 48.8$\pm$0.6 & 49.8$\pm$0.6 & 77.6$\pm$0.2 \\
& Real-IAD Variety & 160 & 1 & 84.3$\pm$0.1 & 96.9$\pm$0.0 & 94.2$\pm$0.0 & 92.5$\pm$0.1 & 48.9$\pm$0.4 & 49.9$\pm$0.4 & 77.8$\pm$0.2 \\
\rowcolor[gray]{0.95} \multicolumn{4}{c}{$\Delta \downarrow$} & 1.4 & 0.3 & 0.2 & 1.4 & 1.2 & 1.0 & 3.6 \\ \midrule

% \multirow{6}{*}{AdaptCLIP~\cite{gao2025adaptclip}} 
% & Real-IAD & 30 & 2 & 83.0$\pm$0.1 & 81.6$\pm$0.2 & 74.4$\pm$0.0 & 97.3$\pm$0.0 & 37.7$\pm$0.2 & 43.6$\pm$0.2 & 88.6$\pm$0.1 \\
% & Real-IAD Variety S1 & 30 & 2 & 85.5$\pm$0.1 & 97.3$\pm$0.0 & 94.6$\pm$0.0 & 91.4$\pm$0.0 & 51.2$\pm$0.1 & 51.6$\pm$0.1 & 75.5$\pm$0.2 \\
% & Real-IAD Variety S2 & 60 & 2 & 85.3$\pm$0.2 & 97.1$\pm$0.0 & 94.4$\pm$0.0 & 91.7$\pm$0.0 & 50.4$\pm$0.4 & 50.8$\pm$0.4 & 76.8$\pm$0.1 \\
% & Real-IAD Variety S3 & 100 & 2 & 86.1$\pm$0.1 & 97.2$\pm$0.0 & 94.5$\pm$0.0 & 92.5$\pm$0.0 & 51.0$\pm$0.1 & 51.5$\pm$0.1 & 78.6$\pm$0.1 \\
% & Real-IAD Variety & 160 & 2 & 86.4$\pm$0.1 & 97.3$\pm$0.1 & 94.5$\pm$0.0 & 92.8$\pm$0.0 & 50.8$\pm$0.1 & 51.5$\pm$0.2 & 78.7$\pm$0.1 \\
% \rowcolor[gray]{0.95} \multicolumn{4}{c}{$\Delta \downarrow$} & 1.1 & 0.2 & 0.2 & 1.4 & 0.8 & 0.8 & 3.2 \\ \midrule

\multirow{6}{*}{AdaptCLIP~\cite{gao2025adaptclip}} 
& Real-IAD & 30 & 4 & 84.0$\pm$0.2 & 82.6$\pm$0.2 & 75.2$\pm$0.2 & 97.4$\pm$0.0 & 38.8$\pm$0.1 & 44.4$\pm$0.1 & 89.1$\pm$0.1 \\
& Real-IAD Variety S1 & 30 & 4 & 87.4$\pm$0.2 & 97.6$\pm$0.1 & 94.8$\pm$0.1 & 91.7$\pm$0.0 & 52.8$\pm$0.2 & 52.8$\pm$0.1 & 76.4$\pm$0.1 \\
& Real-IAD Variety S2 & 60 & 4 & 87.3$\pm$0.1 & 97.5$\pm$0.0 & 94.6$\pm$0.0 & 92.0$\pm$0.0 & 52.1$\pm$0.2 & 52.2$\pm$0.2 & 77.7$\pm$0.0 \\
& Real-IAD Variety S3 & 100 & 4 & 88.0$\pm$0.1 & 97.6$\pm$0.0 & 94.7$\pm$0.0 & 92.8$\pm$0.0 & 52.6$\pm$0.2 & 52.9$\pm$0.2 & 79.4$\pm$0.0 \\
& Real-IAD Variety & 160 & 4 & 88.1$\pm$0.1 & 97.7$\pm$0.0 & 94.7$\pm$0.0 & 93.0$\pm$0.0 & 52.5$\pm$0.2 & 52.9$\pm$0.2 & 79.4$\pm$0.1 \\
\rowcolor[gray]{0.95} \multicolumn{4}{c}{$\Delta \downarrow$} & 0.8 & 0.2 & 0.2 & 1.3 & 0.7 & 0.7 & 3.0 \\ 
\bottomrule
\end{tabular}
}
%\endgroup % 
\end{table}

\begin{figure*}[t]
    \centering
    \includegraphics[width=1.0\linewidth]{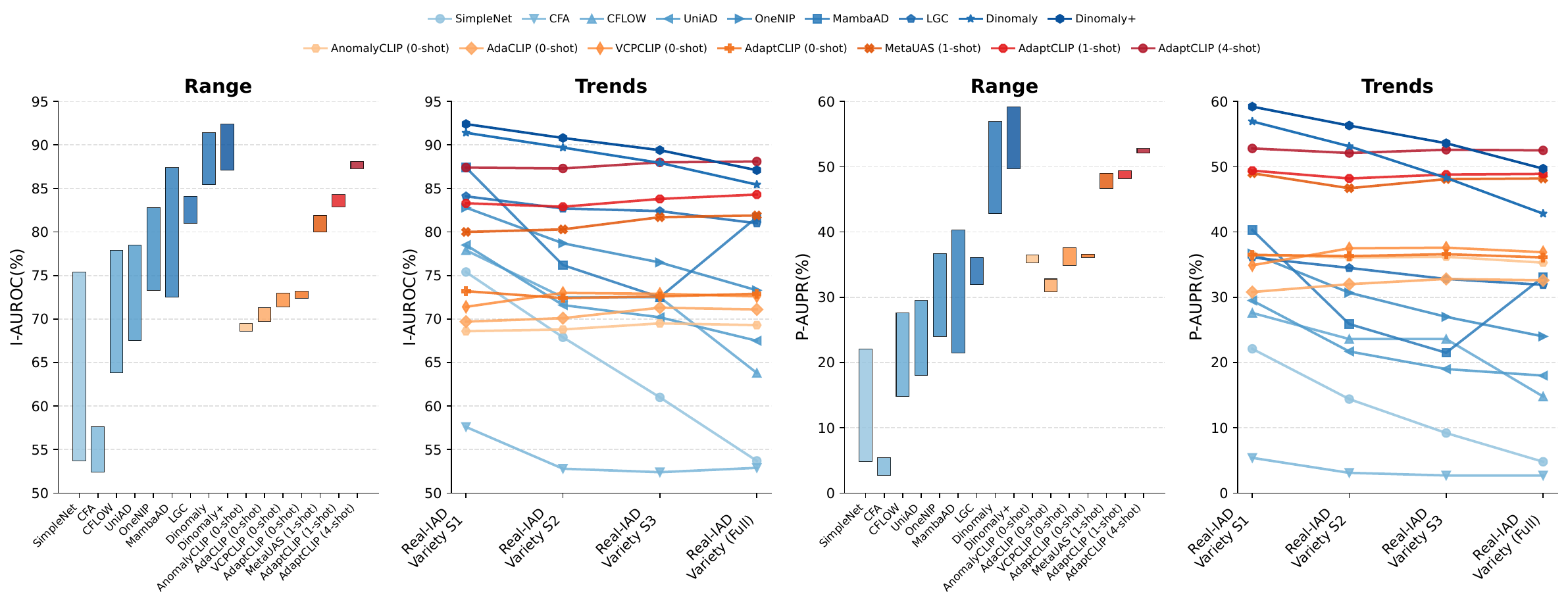}
    \caption{
        Performance trends of I-AUROC and P-AUPR metrics with increasing categories for MUAD, ZSAD and FSAD methods.
    }
    \label{fig:incre}
\end{figure*}

\subsection{Benchmark on Zero-Shot and Few-Shot AD}

\noindent\textbf{Experimental Setting}.
Zero-Shot Anomaly Detection (ZSAD) requires models trained on auxiliary datasets to identify anomalies in novel object categories without class-specific training data.
This setting is particularly valuable because obtaining comprehensive coverage of all object classes and anomaly types is impractical in real-world scenarios.
ZSAD aims to enhance model robustness and adaptability by facilitating knowledge transfer from known to unknown categories, thereby enabling effective detection of novel anomalies across diverse and dynamic environments.
To further evaluate model adaptability in open-world scenarios, we introduce a Few-Shot Anomaly Detection (FSAD) setting alongside ZSAD evaluation.
In FSAD, a limited number of normal images ($k \in \{1,4\}$) from the target class serve as visual prompts during inference, with their detection results integrated with ZSAD outputs to enhance final anomaly detection performance. \gbb{To ensure statistical significance, performance for all FSAD models is averaged across three runs with different random seeds, reporting both mean and standard deviation.}

\noindent\textbf{Competitive Methods}.
We evaluate AnomalyCLIP~\cite{iclr2024anomalyclip}, AdaCLIP~\cite{eccv24adaclip}, VCP-CLIP~\cite{eccv24vcpclip}, and AdaptCLIP~\cite{gao2025adaptclip} on Real-IAD Variety in \gbb{a} zero-shot manner, conducting comprehensive comparisons with Real-IAD~\cite{realiad}.
Given that Real-IAD Variety comprises 160 novel categories, it serves as an exceptionally challenging benchmark for assessing ZSAD model robustness and cross-category generalization in complex real-world scenarios.
Notably, AdaptCLIP~\cite{gao2025adaptclip} supports both zero- and few-shot generalization across domains. Therefore, we employ it to evaluate FSAD performance on Real-IAD Variety. In addition, we also evaluate the one-shot pure-vision MetaUAS~\cite{metauas}.

\noindent\textbf{Results and Analysis under ZSAD and FSAD}.
Table~\ref{tab:zsad_multicatergories} presents ZSAD and FSAD results on Real-IAD and Real-IAD Variety, with performance trends visualized in Fig.~\ref{fig:incre}.

Some key observations are as follows:
(1) ZSAD models demonstrate remarkable stability and robustness, while traditional MUAD methods exhibit significant performance degradation as category complexity increases. This fundamental difference is quantitatively highlighted by low I-ROC $\Delta$ values (\eg, 0.9 for AnomalyCLIP, 0.4 for AdaptCLIP) across increasing category counts (30 to 160). This suggests that VLM-based ZSAD approaches, which leverage external knowledge from vast pre-training data rather than modeling intrinsic training set distributions, are inherently less sensitive to data distribution shifts induced by category scale-up.
(2) FSAD yields substantial performance gains by \gbb{leveraging class-specific prompt information}. Specifically, using only 4 normal images (4-Shot), AdaptCLIP's performance on Real-IAD Variety (160 classes) improves significantly, with I-AUROC rising from 72.9\% (ZSAD) to 88.1\% (4-Shot).
This improvement confirms the high value of incorporating even few-shot normal visual prompts to rapidly refine generic VLM knowledge for precise, target-specific anomaly discrimination. The consistent improvement from 1-Shot to 4-Shot further validates the effectiveness of this adaptation mechanism.
(3) Few-shot AdaptCLIP (4-Shot) achieves the best overall performance across all benchmarks. On the most challenging 160-class Real-IAD Variety, 4-Shot AdaptCLIP reaches I-ROC of 88.1\%, which is highly competitive with the best multi-class Dinomaly~\cite{guo2025dinomaly} (85.4\%) that utilizes full normal training data. Crucially, delta analysis reveals that while few-shot adaptation boosts absolute performance, it incurs slightly higher delta for pixel-level metrics, suggesting that local adaptation mechanisms, while powerful, exhibit greater sensitivity to category complexity than global ZSAD paradigms.

\gbb{ZSAD and FSAD leverage semantic anchoring via textual prompts or visual representation with normal images prompts. Unlike MUAD, these methods utilize a pre-aligned latent space where linguistic concepts of anomalies (\eg, ``damaged'', ``defect'') provide a unified representation of common anomalous (normal visual) features. Because they rely on frozen VLMs and semantic reasoning rather than statistical learning on specific datasets, they require no category-specific fine-tuning. Consequently, their generalization capability is fundamentally decoupled from the number of test categories. MUAD models may suffer from feature       ``pollution'' or capacity constraints as more classes are added. In contrast, ZSAD models identify defects based on pre-trained semantic alignment rather than statistical novelty relative to a specific training set.}
To further unlock the potential of these models, particularly in the complex cross-category setting of Real-IAD Variety, future research should focus on developing enhanced methods that effectively utilize detailed textual and semantic information, bridging the gap between high-level semantic understanding and fine-grained localized visual anomaly detection.

\subsection{Benchmark on Multi-View AD}

\noindent\textbf{Experimental Setting}.
Multi-View Anomaly Detection (MVAD) originates from high-precision demands in industrial quality inspection.
It aims to achieve detection capabilities unattainable from single-view images by integrating multiple perspective images of the same sample, thereby \gbb{detecting defects that may be occluded or invisible in certain viewpoints but remain discernible in others.}
Additionally, this setting leverages multi-view information and consistency constraints to further enhance model performance.

\begin{table}[ht]
    \centering
    \gbb{
    \caption{MVAD~\cite{mvad} performance comparison of Real-IAD Variety with increasing categories.}
    \label{tab:mvad}
    \resizebox{\columnwidth}{!}{ 
        \begin{tabular}{lccccccc} 
            \toprule
            \textbf{Model/Dataset} & \textbf{I-ROC} & \textbf{I-PR} & \textbf{I-F1$_{max}$} & \textbf{P-ROC} & \textbf{P-PR} & \textbf{P-F1$_{max}$} & \textbf{P-PRO} \\ 
            \midrule
            Real-IAD-30c         &86.6	&84.8	&77.2	&97.9	&30.3	&36.8	&91.2 \\
            Real-IAD Variety S1  & 85.4 & 97.0 & 94.8 & 90.7 & 36.8 & 39.7 & 72.4 \\
            Real-IAD Variety S2  & 81.4 & 96.2 & 94.1 & 90.3 & 31.8 & 36.2 & 71.0 \\
            Real-IAD Variety S3  & 80.3 & 95.8 & 94.0 & 90.5 & 28.8 & 33.9 & 70.5 \\
            Real-IAD Variety     & 77.1 & 95.1 & 93.6 & 89.5 & 24.5 & 30.2 & 67.7 \\
            \bottomrule
        \end{tabular}
    } }
\end{table}

\noindent\textbf{Results and Analysis under MVAD}.
Table~\ref{tab:mvad} presents quantitative results of MVAD~\cite{mvad} on Real-IAD Variety.
As category count increases, model performance gradually decreases, consistent with trends observed in MUAD settings.
In practical production line applications, multi-view imaging is essential to reduce false negative rates.
Nevertheless, current metric results remain relatively modest, presenting a significant challenge for future research to address scalability and robustness in large-scale multi-view anomaly detection scenarios.

\subsection{Potential for Foundational AD Models}

\begin{table}[ht]
\centering
\gbb{
\caption{
Few-shot anomaly detection results on MVTec and VisA. We compare the ADPretrain model (FeatureNorm) pre-trained on Real-IAD versus our Real-IAD Variety.}
\label{tab:fsad-adpretrain}
\small
\resizebox{1.0\columnwidth}{!}{
\begin{tabular}{cccccccccc}
\toprule
\textbf{Test Set} & \textbf{Train Set} & \textbf{Shot} & \textbf{I-ROC} & \textbf{I-PR} & \textbf{I-F1$_{max}$} & \textbf{P-ROC} & \textbf{P-PR} & \textbf{P-F1$_{max}$} & \textbf{P-PRO} \\ 
\midrule
\multirow{8}{*}{MVTec} & \multirow{4}{*}{Real-IAD} & 1 & 87.9 & 93.8 & 90.2 & 96.7 & 48.1 & 53.4 & 91.4 \\
 &  & 2 & 95.6 & 97.0 & 95.3 & 97.3 & 50.2 & 55.2 & 92.3 \\
 &  & 4 & 97.4 & 98.3 & 96.6 & 97.4 & 50.6 & 55.6 & 92.5 \\
 &  & 8 & 97.8 & 98.6 & 96.8 & 97.5 & 51.5 & 56.0 & 92.8 \\
 \cmidrule{2-10}
 & \multirow{4}{*}{\shortstack[l]{Real-IAD Variety}} & 1 & 90.7 & 93.8 & 93.1 & 97.3 & 54.1 & 56.6 & 92.0 \\
 &  & 2 & 96.0 & 97.5 & 95.8 & 97.7 & 56.1 & 58.1 & 92.9 \\
 &  & 4 & 97.7 & 99.0 & 97.2 & 98.0 & 56.9 & 58.6 & 93.6 \\
 &  & 8 & 98.1 & 99.1 & 97.5 & 98.1 & 57.4 & 59.1 & 93.7 \\
\midrule
\multirow{8}{*}{ViSA} & \multirow{4}{*}{Real-IAD} & 1 & 61.4 & 65.7 & 73.2 & 96.4 & 19.2 & 28.0 & 81.0 \\
 &  & 2 & 68.8 & 75.1 & 74.1 & 97.0 & 25.1 & 33.5 & 83.7 \\
 &  & 4 & 75.2 & 81.5 & 77.2 & 97.6 & 28.8 & 37.3 & 85.5 \\
 &  & 8 & 82.6 & 84.1 & 80.4 & 98.0 & 31.7 & 40.9 & 87.1 \\
 \cmidrule{2-10}
 & \multirow{4}{*}{\shortstack[l]{Real-IAD Variety}} & 1 & 58.6 & 62.4 & 73.1 & 96.8 & 20.8 & 30.7 & 83.1 \\
 &  & 2 & 70.4 & 74.6 & 75.1 & 97.6 & 30.0 & 37.2 & 86.5 \\
 &  & 4 & 88.1 & 90.0 & 84.8 & 98.0 & 34.4 & 40.5 & 88.5 \\
 &  & 8 & 93.4 & 94.4 & 89.6 & 98.3 & 36.1 & 42.1 & 89.7 \\
\bottomrule
\end{tabular}}
}
\end{table}

\noindent\textbf{Experimental Setting}.
\gbb{To demonstrate the generalization of foundational AD models pre-trained large-scale Real-IAD Variety dataset. We employ ADPretrain~\cite{adpretrain} framework (DINOv2-base backbone) and compare the effects of pre-training on the original Real-IAD and our proposed Real-IAD Variety. Then, we evaluate few-shot (1, 2, 4, 8-shot) generalization on the independent MVTec and VisA benchmarks.}

\noindent\textbf{Few-shot Generalization}.
\gbb{As shown in Table~\ref{tab:fsad-adpretrain}, the few-shot performance of ADPretrain model pre-trained on Real-IAD Variety consistently outperforms the baseline across all settings. Notably, the performance gains are significantly more pronounced on the VisA dataset, which is widely recognized as a more challenging benchmark due to its complex object structures and subtle defects. For instance, in the 4-shot setting on VisA, pre-training on Real-IAD Variety improves I-ROC from 75.2\% to 88.1\% (+12.9\%). We observe a slight fluctuation in the 1-shot ViSA setting, which may be attributed to the high sensitivity of extremely low-shot sampling in certain complex categories.
These results empirically demonstrate that the increased category diversity and broader industrial scope of Real-IAD Variety are crucial for enhancing generalization on foundational AD models. By providing a superior foundation for feature alignment, our dataset helps foundational models learn robust visual primitives that generalize far better to challenging unseen domains.}

\section{Future Work}

\noindent\textbf{Foundation Models for Anomaly Detection: Scaling Toward a New Paradigm.}
Foundation models have fundamentally transformed how diverse tasks are addressed through unified architectures, ushering in a new era of model design~\cite{zhao2023survey,zhang2024vision,du2022survey}.
A critical enabler of this transformation is the availability of large-scale, high-quality datasets, which are essential for training versatile and generalizable models across various domains.
In the context of Industrial Anomaly Detection (IAD), the development of foundation models is heavily dependent on comprehensive vision-language annotations.
These annotations are indispensable not only for accurate anomaly detection but also for providing contextual understanding that enables models to reason about anomalies, thereby enhancing their generalization capabilities across diverse industrial scenarios.

\noindent\textbf{Multimodal Anomaly Detection with Large-Scale Diverse Data.}
The introduction of Real-IAD Variety opens several promising research directions for advancing anomaly detection.
First, the development of robust generalized anomaly detection models represents a critical avenue. 
Future research should explore training strategies that effectively integrate these textual insights with visual features, \eg, MMAD~\cite{mmad}, thereby enhancing the robustness and adaptability of anomaly detection systems.
Second, addressing logical anomalies in image-based anomaly detection remains an important challenge. Current methods often struggle with logical anomalies, \gbb{where anomalies are defined by misplaced components or incorrect counts rather than distinctive surface textures. Our ZSAD/FSAD results suggest that while VLMs excel at semantic alignment, bridging the gap between high-level reasoning and fine-grained spatial logic remains a frontier for foundational IAD research.}
Our preliminary ZSAD and FSAD experiments have revealed performance limitations in this regard. To address these constraints, we propose enabling text encoder fine-tuning to enhance model adaptability and contextual reasoning capabilities.
Future research should focus on integrating textual descriptions into existing anomaly detection frameworks, enabling models to reason about logical relationships and contextual cues beyond purely visual patterns.

\section{Conclusion}\label{sec:conclusion}

Industrial Anomaly Detection (IAD) is undergoing a fundamental transformation with the emergence of large-scale models capable of operating in unified and zero-/few-shot settings.
This work makes a substantial contribution to the IAD dataset landscape by introducing Real-IAD Variety, a large-scale benchmark that not only expands the scope of available data but also establishes rigorous evaluation protocols for assessing state-of-the-art (SOTA) IAD methodologies.
Real-IAD Variety, characterized by its unprecedented scale and diversity, redefines the benchmark standards for IAD datasets. It encompasses 160 categories spanning 28 industries, 24 material types, 22 color variations and \gbb{27 defect types} with \gbb{198,950} images, effectively overcoming the limitations of previous datasets that were constrained by narrow category ranges and limited scenario representation.

Our comprehensive experimental analysis on Real-IAD Variety reveals several critical findings:
(1) Most multi-class unsupervised anomaly detection methods experience significant performance degradation as the category count increases, with I-ROC typically declining by 10\%--20\% when scaling from 30 to 160 categories;
(2) Zero-shot and few-shot approaches demonstrate remarkable resilience to category scale-up, exhibiting minimal performance fluctuation. This suggests their superior scalability and effectiveness in handling diverse industrial scenarios;
(3) Vision-language models (VLMs) leveraging external pre-trained knowledge exhibit fundamentally different scalability characteristics compared to traditional unsupervised methods, highlighting the pivotal role of multimodal learning in the next generation of IAD systems.
Real-IAD Variety and its associated benchmarks provide a solid foundation for future research and innovation, particularly in multi-class, multi-view and zero-/few-shot anomaly detection paradigms.
We anticipate that this benchmark will facilitate the development of models with enhanced generalization capabilities, ultimately improving the reliability and efficiency of anomaly detection systems in real-world industrial applications.

\section*{Acknowledgements}

This work was partially supported by the National Natural Science Foundation of China (Grant No. 62171139).

{
    \small
    \bibliographystyle{ieeenat_fullname}
    \bibliography{main}
}

\clearpage
\renewcommand\thefigure{A\arabic{figure}}
\renewcommand\thetable{A\arabic{table}}  
\renewcommand\theequation{A\arabic{equation}}
\setcounter{equation}{0}
\setcounter{table}{0}
\setcounter{figure}{0}
\appendix\label{app}

This supplement provides extended taxonomies (industry, material, and color) and dataset statistics including image resolutions (Fig.~\ref{fig:imageresolution}) and class-wise distributions (Fig.~\ref{fig:imageresolution}) for Real-IAD Variety.

\noindent\textbf{1. Industry Groups} 

% --- C1 ---
\textbf{c1: Electrical Manufacturing} \\ 
%\hspace*{3em} 
$\left\{ \begin{tabular}{p{0.9\columnwidth}
}
    s1: Manufacturing of Power Electronics Components \\
    s2: Manufacturing of Distribution Switch Control Equipment \\
    s3: Manufacturing of Wires and Cables \\
    s4: Manufacturing of Other Power Distribution and Control Equipment \\
    s5: Battery Manufacturing \\
    s6: Motor Manufacturing
\end{tabular} \right.$ %\\[0.8em]

% --- C2 ---
\textbf{c2: Transport Manufacturing} \\
%\hspace*{3em} 
$\left\{ \begin{tabular}{p{0.9\columnwidth}}
    s7: Manufacturing of Distribution Switch Control Equipment \\
    s8: Plastic Toy Manufacturing
\end{tabular} \right.$ %\\[0.8em]

% --- C3 ---
\textbf{c3: Cultural Products Manufacturing} \\
%\hspace*{3em} 
$\left\{ \begin{tabular}{p{0.9\columnwidth}}
    s9: Stationery Manufacturing \\
    s10: Manufacturing of Daily Plastic Products \\
    s11: Metal Tool Manufacturing
\end{tabular} \right.$ %\\[0.8em]

% --- C4 ---
\textbf{c4: Metal Manufacturing} \\
%\hspace*{3em} 
$\left\{ \begin{tabular}{p{0.9\columnwidth}}
    s12: Manufacturing of Metal Accessories for Construction and Furniture \\
    s13: Manufacturing of Decorative Metal Products
\end{tabular} \right.$ %\\[0.8em]

% --- C5 ---
\textbf{c5: General Manufacturing} \\
%\hspace*{3em} 
$\left\{ \begin{tabular}{p{0.9\columnwidth}}
    s14: Manufacturing of Safety and Fire Protection Metal Products \\
    s15: Manufacturing of Daily Miscellaneous Products \\
    s16: General Spare Parts Manufacturing
\end{tabular} \right.$ %\\[0.8em]

% --- C6 ---

\textbf{c6: Electronics Manufacturing} \\
%\hspace*{3em} 
$\left\{ \begin{tabular}{p{0.9\columnwidth}}
    s17: Manufacturing of Resistors Capacitors and Inductors \\
    s18: Manufacturing of Sensors \\
    s19: Manufacturing of Acoustics Devices and Parts \\
    s20: Manufacturing of Semiconductor Lighting Devices \\
    s21: Electronic Circuit Manufacturing
\end{tabular} \right.$ %\\[0.8em]

% --- C7 ---
\textbf{c7: Rubber and Plastic Products Manufacturing} \\
%\hspace*{3em} 
$\left\{ \begin{tabular}{p{0.9\columnwidth}}
    s22: Manufacturing of Plastic Parts and Other Plastic Products \\
    s23: Rubber Parts Manufacturing \\
    s24: Daily Plastic Products Manufacturing \\
    s25: Daily Miscellaneous Products Manufacturing
\end{tabular} \right.$ %\\[0.8em]

% --- C8 ---
\textbf{c8: Other} \\
%\hspace*{3em} 
$\left\{ \begin{tabular}{p{0.9\columnwidth}}
    s26: Manufacturing of Clothing Supplies \\
    s27: Manufacturing of Automotive Parts and Accessories \\
    s28: Clock and Timing Instrument Manufacturing
\end{tabular} \right.$ %\\[0.8em]

\vspace{5pt}
\noindent\textbf{2. Material Types} %\\[0.8em]

% --- u1: Metal ---
\textbf{u1: Metal} \\
$\left\{ \begin{tabular}{p{0.9\columnwidth}}
    v1: Aluminum Alloy \\
    v2: Zinc Alloy \\
    v3: Iron Oxide \\
    v4: Steel \\
    v5: Lithium \\
    v6: Copper
\end{tabular} \right.$ 

% --- u2: Plastic ---
\textbf{u2: Plastic} \\
$\left\{ \begin{tabular}{p{0.9\columnwidth}}
    v7: ABS \\
    v8: PC \\
    v9: PA \\
    v10: PPS \\
    v11: PET \\
    v12: PS \\
    v13: PP \\
    v14: POM \\
    v15: PBT \\
    v16: PVC \\
    v17: Acrylic \\
    v18: PE
\end{tabular} \right.$ %\\[0.8em]

% --- u3: Other ---
\textbf{u3: Other} \\
$\left\{ \begin{tabular}{p{0.9\columnwidth}}
    v19: Silicon \\
    v20: Carbon \\
    v21: Rubber \\
    v22: Ceramics \\
    v23: Wood \\
    v24: Glass cup epoxy resin
\end{tabular} \right.$ %\\[0.8em]

\vspace{5pt}
\noindent\textbf{3. Color Types} 

% --- a1 ---
\textbf{a1: Chroma} \\
$\left\{ \begin{tabular}{p{0.9\columnwidth}}
    b1: Gray \\
    b2: Pink \\
    b3: Green \\
    b4: Purple \\
    b5: Blue \\
    b6: Red \\
    b7: Yellow \\
    b8: White \\
    b9: Gold \\
    b10: Brown
\end{tabular} \right.$ %\\[0.8em]

% --- a2 ---
\textbf{a2: Mixer Color} \\
$\left\{ \begin{tabular}{p{0.9\columnwidth}}
    b11: Green + Silver \\
    b12: Black + silver \\
    b13: Black + Yellow \\
    b14: Yellow + Silver \\
    b15: Blue + white \\
    b16: Black + Gold \\
    b17: Silver + blue \\
    b18: Black + white \\
    b19: Black + blue \\
    b20: Red + silver \\
    b21: Silver + gold \\
    b22: White + silver
\end{tabular} \right.$ %\\[0.8em]

% --- a3 (单项) ---
\textbf{a3: Silver} %\\[0.8em]

% --- a4 (单项) ---
\textbf{a4: Black}

\begin{figure*}[ht]
\centering
\includegraphics[width=0.9\linewidth]{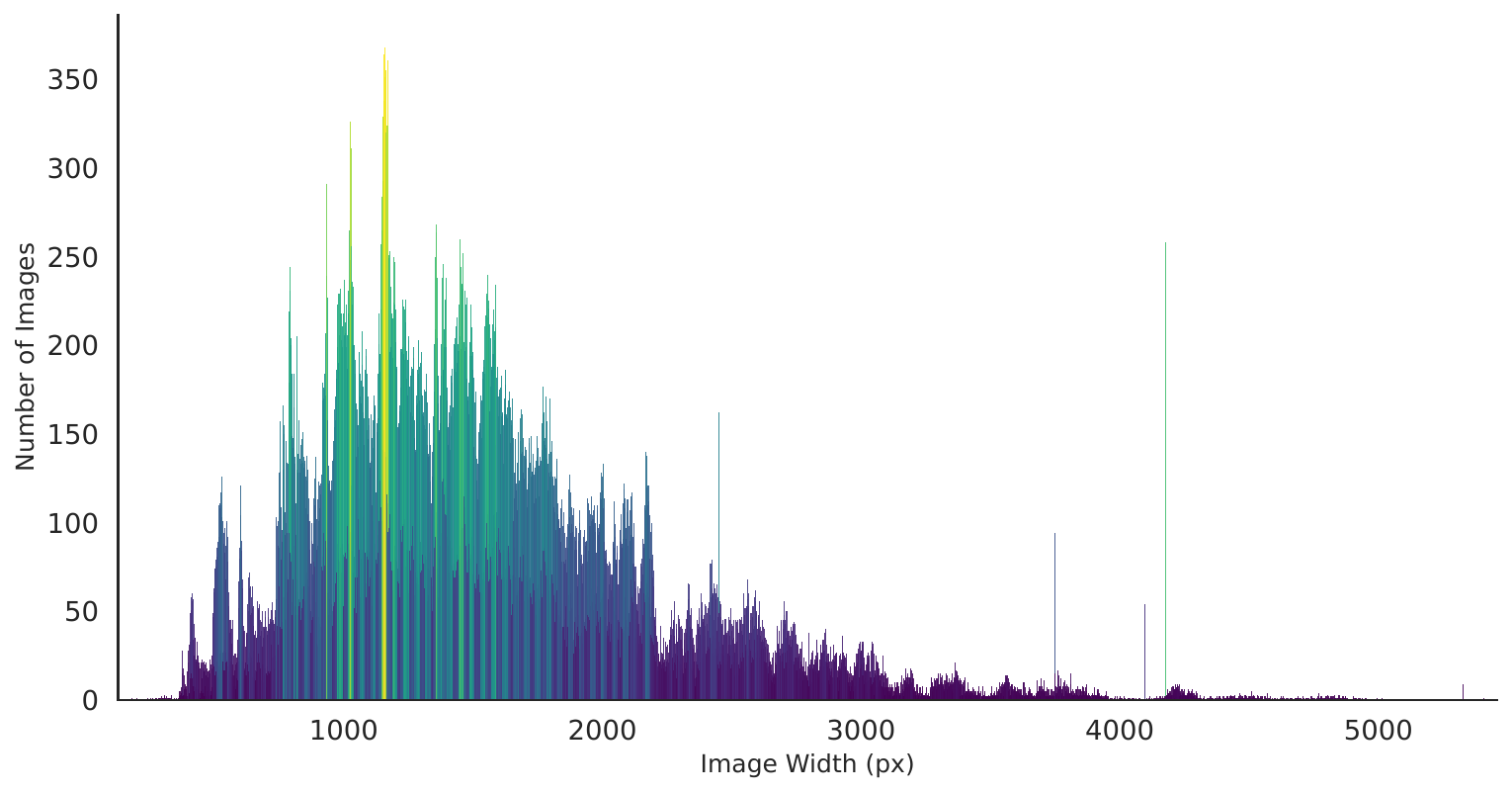}
\includegraphics[width=0.9\linewidth]{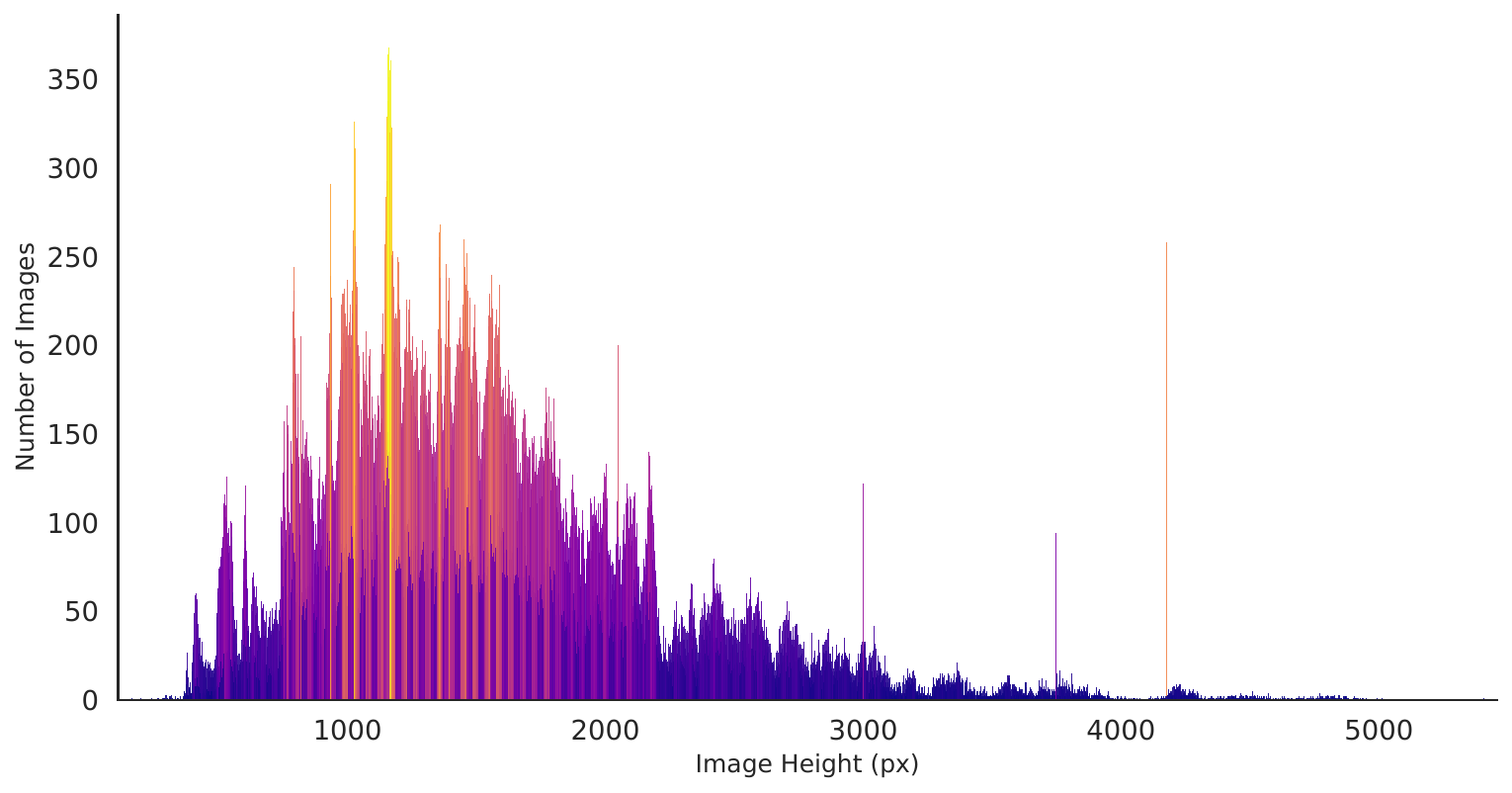}
\caption{The distributions of image height and width in our Real-IAD-Variety.}
\label{fig:imageresolution}
\end{figure*}

\begin{figure*}[ht]
\centering
\includegraphics[width=0.9\linewidth]{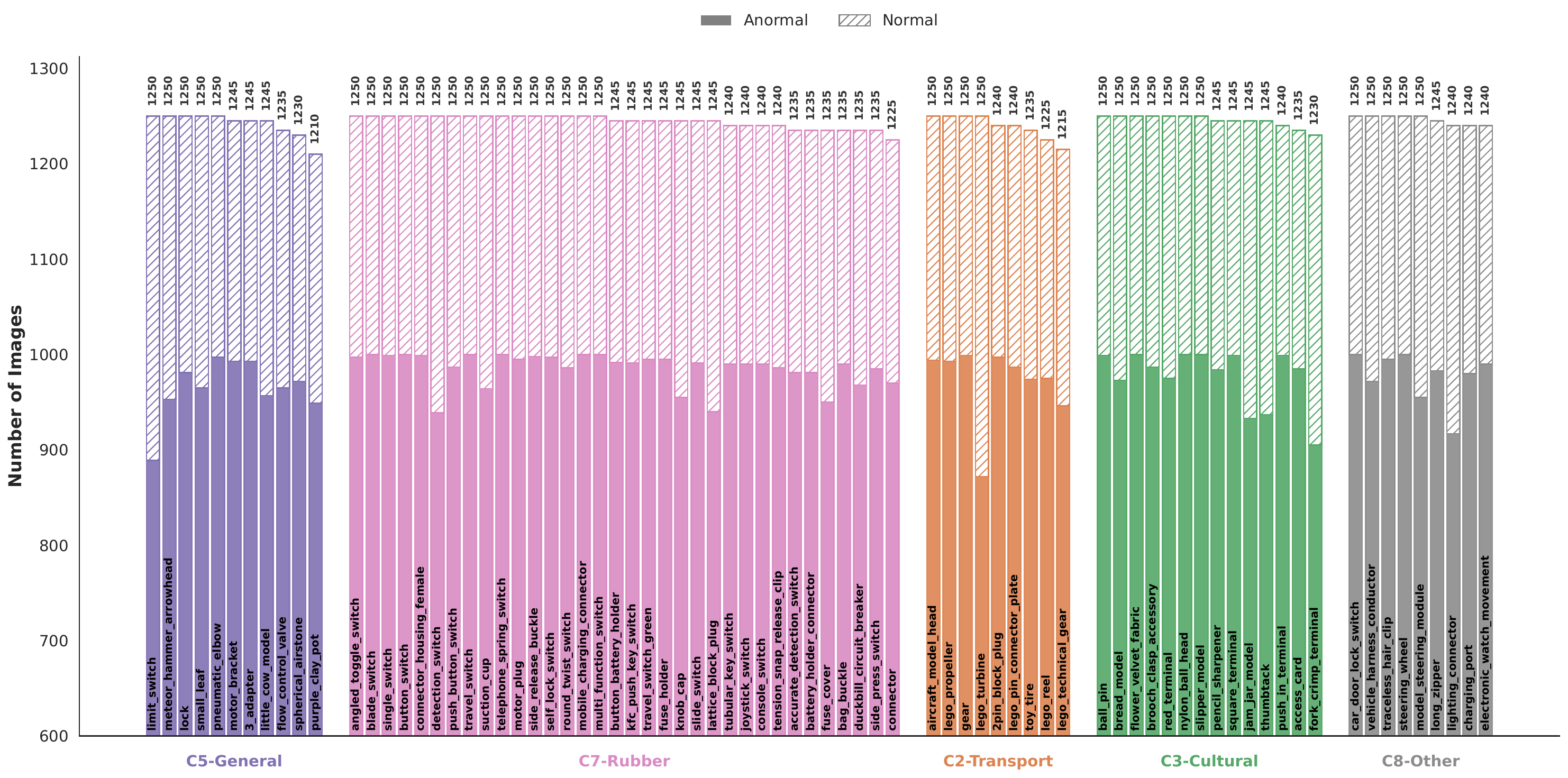}
\includegraphics[width=0.9\linewidth]{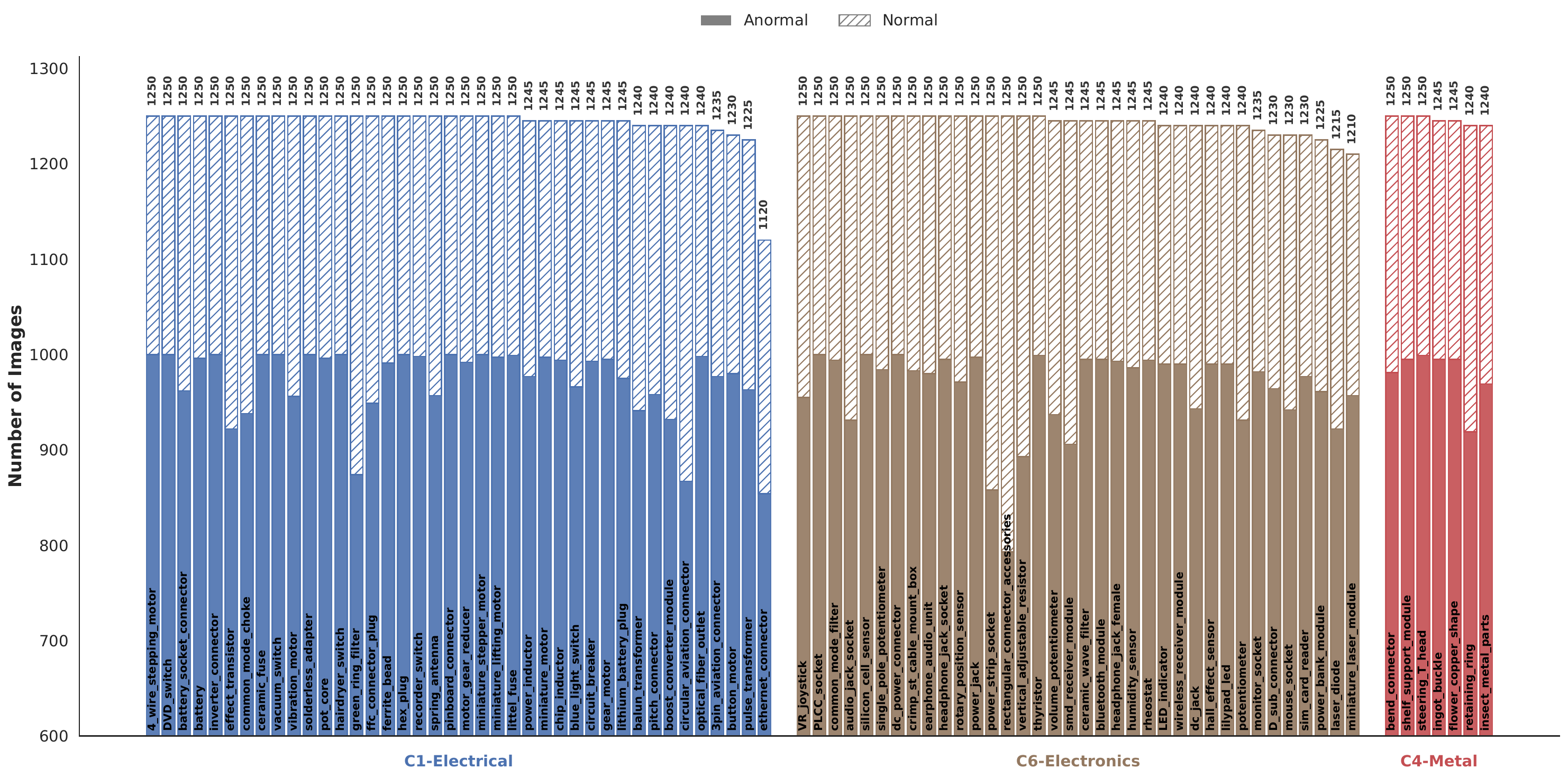}
\caption{The class-wise distribution of normal and abnormal images in our Real-IAD Variety dataset.}
\label{fig:imageresolution}
\end{figure*}

\end{document}